\pdfoutput=1

\documentclass[11pt]{article}

\usepackage[]{ACL2023}

\usepackage{times}
\usepackage{latexsym}

\usepackage[T1]{fontenc}

\usepackage[utf8]{inputenc}

\usepackage{microtype}

\usepackage{inconsolata}

\pdfoutput=1
\usepackage{graphics}
\usepackage{graphicx}
\usepackage{booktabs}
\usepackage{multirow, makecell}
\usepackage[disable]{todonotes}
\usepackage{amsmath}
\usepackage[normalem]{ulem}
\usepackage{caption}
\usepackage{subcaption}
\usepackage[greek,english]{babel}
\usepackage{mathtools}
\usepackage{pdflscape}

\newcommand{\block}[1]{%
  \raisebox{\dimexpr(\fontcharht\font`X-1em)/2}{\rule{1em}{#1\dimexpr1em/8}}%
}
\DeclareUnicodeCharacter{2581}{\block{1}}
\DeclareUnicodeCharacter{2582}{\block{2}}
\DeclareUnicodeCharacter{2583}{\block{3}}
\DeclareUnicodeCharacter{2584}{\block{4}}
\DeclareUnicodeCharacter{2585}{\block{5}}
\DeclareUnicodeCharacter{2586}{\block{6}}
\DeclareUnicodeCharacter{2587}{\block{7}}
\DeclareUnicodeCharacter{2588}{\block{8}}

\newcommand{\va}[0]{\emph{vocabulary allocation}}
\newcommand{\vo}[0]{\emph{vocabulary overlap}}

\DeclareMathOperator{\rank}{rank}

\definecolor{bottlegreen}{rgb}{0.0,0.42,0.31}
\definecolor{babypink}{rgb}{0.83,0.42,0.54}
\definecolor{royalpurple}{rgb}{.471,.318,.663}

\newcommand{\draftcomment}[3]{{\textcolor{#3}{[#1]#2}}}
\renewcommand{\draftcomment}[3]{}  

\newcommand{\tomasz}[1]{\draftcomment{#1}{\textsc{tomasz}}{bottlegreen}}
\newcommand{\tomaszrep}[2]{\tomasz{\sout{#1} #2}}

\newcommand{\jirka}[1]{\draftcomment{#1}{\textsc{jirka}}{babypink}}

\newcommand{\david}[1]{\draftcomment{#1}{\textsc{david}}{orange}}

\title{Tokenization Impacts Multilingual Language Modeling: \\
Assessing Vocabulary Allocation and Overlap Across Languages}

\author{Tomasz Limisiewicz \and  Jiří Balhar \and David Mareček\\
            Institute of Formal and Applied Linguistics, Faculty of Mathematics and Physics \\
            Charles University, Prague, Czech Republic \\
  \texttt{\{limisiewicz, marecek\}@ufal.mff.cuni.cz}
  }

\begin{document}
\maketitle
\begin{abstract}
Multilingual language models have recently gained attention as a promising solution for representing multiple languages in a single model. 
In this paper, we propose new criteria to evaluate the quality of lexical representation and vocabulary overlap observed in sub-word tokenizers.
Our findings show that the overlap of vocabulary across languages can be actually detrimental to certain downstream tasks (POS, dependency tree labeling). In contrast, NER and sentence-level tasks (cross-lingual retrieval, NLI) benefit from sharing vocabulary. We also observe that the coverage of the language-specific tokens in the multilingual vocabulary significantly impacts the word-level tasks.
\tomaszrep{Due to this aspect, the SentencePiece Unigram tokenizer is not the best choice for multilingual language modeling.}{} 
Our study offers a deeper understanding of the role of tokenizers in multilingual language models and guidelines for future model developers to choose the most suitable tokenizer for their specific application before undertaking costly model pre-training.\footnote{The code is available at: \url{github.com/tomlimi/entangled_in_scripts}.}

\end{abstract}

\section{Introduction}
\label{sec:intro}

\pdfoutput=1
\begin{figure}[tb!]
    \centering
    \includegraphics[width=\linewidth]{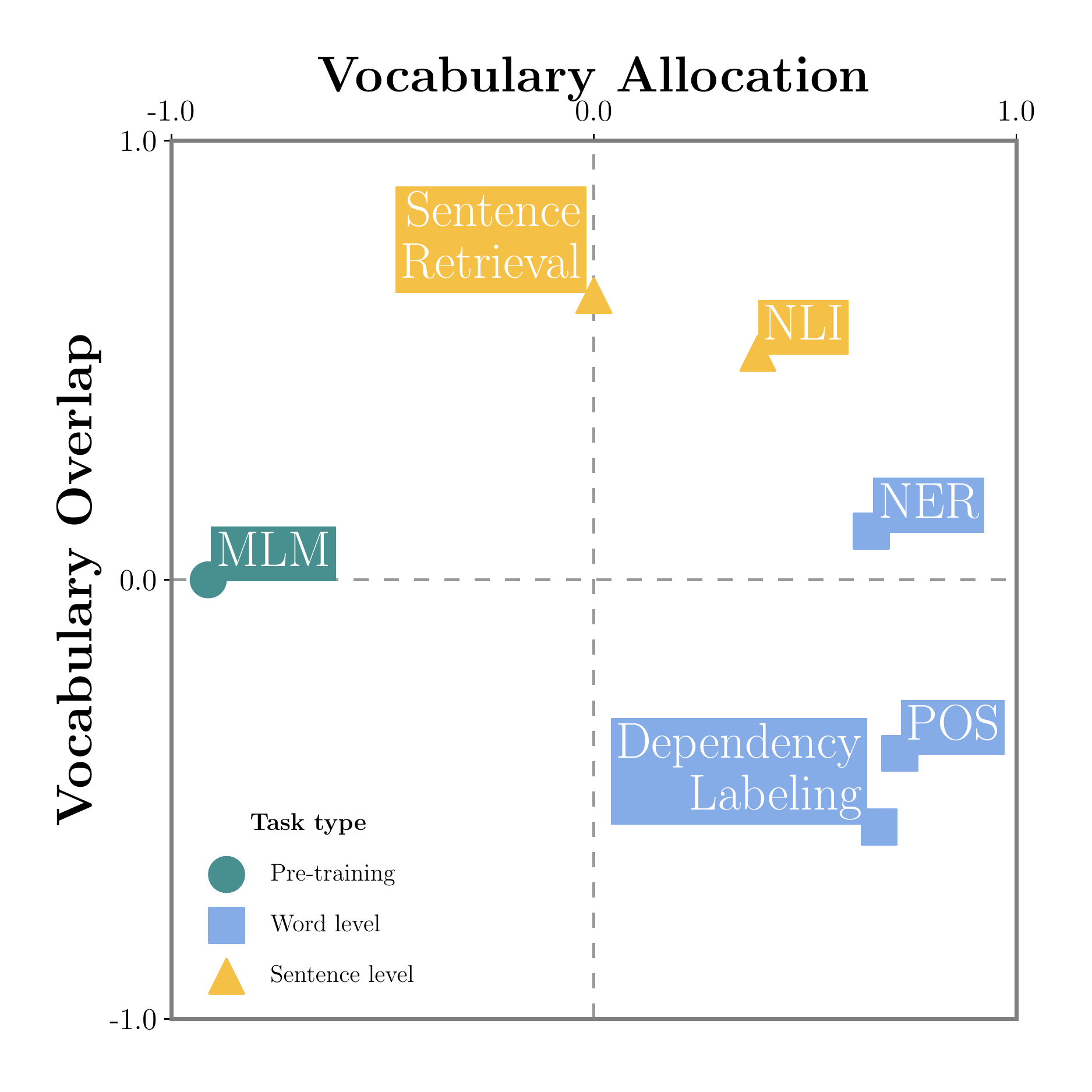}

    \caption{Mapping the impact of \va~and \vo~on language model performance. The location of points corresponds to Spearmnan's correlation between vocabulary measures and the task score (see the details in Tables~\ref{tab:corr_in_lang_20l}~and~\ref{tab:corr_x_lang_6l}). High \vo~benefits NER and sentence-level tasks (NLI, sentence retrieval) and hinders POS and dependency labeling performance. High \va~improves word-level tasks but leads to a decrease in masked language modeling scores. 
    Masked language modeling is measured only in language. Thus it's unaffected by \vo. Analogically, sentence retrieval is solely cross-lingual and unaffected by \va.}
    \label{fig:schwartz}
\end{figure}

Multilingual language models perform surprisingly well in a variety of NLP tasks for diverse languages \cite{devlin_bert_2019, conneau_cross-lingual_2019, conneau_unsupervised_2019}. It has been observed that the representation of the input sequence has a significant effect on their effectiveness \cite{mielke_between_2021}. In the widely used Transformer \cite{vaswani_attention_2017} models achieving state-of-the-art results through diverse tasks, a large fraction of parameters are allocated in the input encoding layer.\footnote{For instance, in XLM-Roberta\textsubscript{Base}, 192M out of 270M parameters are in the input embedding layer (approximately 70\%).
 \jirka{Full size XLM-Roberta has 192M out of 550M (35\%)}
 }
The popular language-independent approach to represent the input texts is to learn a vocabulary of frequently appearing strings that may consist of words or parts of words  \cite{sennrich_neural_2016, song_fast_2021, kudo_sentencepiece_2018}.


In this work, we focus on the characteristics of subword tokenization methods in a multilingual setting.
Our main contribution is the introduction of the methods for measuring whether tokenizers effectively represent meaningful language-specific tokens in the vocabulary (\va) and whether the units they learn are shared across languages (\vo).
We posit the following questions:

\clearpage
\textbf{(Q1) How do sub-word tokenizers differ in \emph{overlap}~and~\emph{allocation} of learned vocabularies?}
To answer this question, we apply the metrics to tokenizers obtained with two widely used algorithms: SentencePiece Unigram LM \cite{kudo_sentencepiece_2018}, and BPE \cite{sennrich_neural_2016}. Furthermore, we propose two methods of learning tokenizers on monolingual corpora and then combining them to allow the tokenization of multilingual texts.

\textbf{(Q2) Which properties of multilingual tokenizers affect the LM's representation quality?} 
We address this question by training small language models utilizing different tokenization methods. We evaluate the models on masked word prediction and a diverse set of downstream tasks: POS, NER tagging, dependency tree labeling, NLI, and cross-lingual sentence retrieval. 

The proposed evaluation scheme  offers a good prediction of language models' performance. Notably, we show that the system results significantly improve when tokenizers allocate more vocabulary units for specific languages. Our investigation shows that this aspect has a bigger influence than the \vo~for word-level tasks (see Figure~\ref{fig:schwartz}).
To the best of our knowledge, the interactions between multilingual \va{}~and \vo{}~have not been investigated in past research.

\section{Multilingual Subword Tokenization}
\label{sec:tokenization}
The majority of the currently deployed models use subword tokenization as a way to pre-process the input texts. The input is represented as a sequence of units from a finite vocabulary, which can be translated into numeric representation by an input embedding layer.

The benefits of subword tokenization are the ability to obtain numeric representation for meaningful words frequently used in the resources and handling less frequent words by splitting them into subwords. The latter property mitigates the problem of out-of-vocabulary (OOV) words by breaking them down into smaller parts (sub-words) already present in the vocabulary. It is crucial in handling multilingual texts, especially in languages with large vocabularies and complex morphology.

In the following section, we describe two widely used algorithms of subword tokenization: 

\subsection{Background: Subword Tokenization}

\textbf{Byte-pair encoding BPE:} \cite{sennrich_neural_2016} is a subword tokenization method that iteratively replaces the most frequent pair of vocabulary units in the input text with a single unit. The process starts with taking unique characters of the training text as the initial vocabulary. Subsequently, we take the most frequent pair of vocabulary units, merge the pair, and add it as a new unit to the vocabulary.  
This process is repeated until a pre-set vocabulary size $N$ is reached.






\textbf{Unigram LM:} \cite{kudo_subword_2018} is the method of obtaining subword vocabulary that was  first introduced as the underlying tokenizer of SentencePiece algorithm \cite{kudo_sentencepiece_2018}. The prerequisite is obtaining an extensive vocabulary, e.g., consisting of all strings present in data with at most, a predefined number of characters.
The expectation-maximization algorithm is used
to estimate the probability of vocabulary units.
After EM convergence, the portion of units with the lowest contribution to the likelihood of the training corpus is removed from the vocabulary.
The procedure is repeated until the pre-set vocabulary size is obtained.



\subsection{Combining Monolingual Tokenizers}
\citet{rust_how_2021} observed that
subword tokenizers trained on monolingual data outperform multilingual ones.
The latter can overrepresent the subwords specific to languages constituting a large portion of the training corpora (e.g., English). Moreover, their vocabulary is less likely to contain morphemes important in modeling low-resource languages and instead prioritizes less meaningful character sequences appearing across languages.

 To alleviate this issue, we suggest utilizing monolingual tokenizers for multilingual tokenization. First, the Unigram LM tokenizers are trained on separate monolingual corpora.
 The tokenizers are then combined to create a tokenizer suitable for multilingual data. We propose two methods for combining monolingual tokenizers:

\textbf{Language-specific Tokenization \textsc{NoOverlap}:} We train Unigram tokenizers for each of $L$ considered languages with the same vocabulary size for each of the languages $\frac{N}{L}$. In multilingual tokenization, we apply the tokenizer for a specific language separately and produce a token with language identification.\footnote{Only the special tokens are shared across languages, e.g., ``<s>'' -- the beginning of a sentence token.} The vocabulary consists of $L$ segments of total size $N$. Naturally, the tokenized texts in different languages will consist of tokens from distinct vocabulary segments. Noticeably, the same character sequence in different languages can be assigned different token ids. 

\textbf{Language-Mixed Tokenization \textsc{TokMix}:} We train Unigram LM tokenizers \david{what is the vocabulary size here?} for each of $L$ languages. Subsequently, we averaged vocabulary unit probabilities across tokenizers, sorted them, and trimmed the vocabulary to the pre-set vocabulary size $N$ keeping the units with the highest 
probability. 
\footnote{To account for possible overlaps between language-specific vocabularies, we set their sizes above $\frac{N}{L}$. It assures that joint vocabulary will have at least $N$ tokens.}

\begin{equation}
    \hat{\theta} = \sum^{L}_{i=1} w_{i} \theta_{i}
\end{equation}
$w_i$ are weights assigned to each language. By default, we set the weights to be uniform and equal to $\frac{1}{L}$. Unlike \textsc{NoOverlap}, the same vocabulary units coming from distinct monolingual tokenizers are merged into one unit with averaged probability.

\subsection{Tokenizer and Model Training Setting}

We initially focused on a group of 6 languages varying both in the script and language family: Arabic, Chinese, Greek, Turkish, Spanish, and English. In subsequent experiments, we extend the method to 20 languages. 

We download $10\%$ of CC corpus available atv \url{https://data.statmt.org/cc-100/}. Following the  methodology in \cite{conneau_cross-lingual_2019}, we subsample each language's data to ensure that the training corpus is well-balanced across languages. An equation defines the sample size $c_l$ for language $l$:

\begin{equation}
    c_{l,\alpha}  = c_{\min} \cdot \left(\frac{|C_l|}{c_{\min}}\right)^\alpha
\end{equation}

Where $c_{\min}$  is the minimal sample size (defined by the smallest language), and $C_l$ is all data available for a language, $\alpha$ is the so-called ``balancing parameter''. In our experiments, we set $c_{\min}$ to 10 M characters, $C_l$ is, e.g., 8.8 B characters for English. We set $\alpha$ to $0.25$, which corresponds to a balancing factor picked for XLM-Roberta \cite{conneau_unsupervised_2019}. The training data for the tokenizer and the model are the same. The vocabulary size $N$ was set to 120,000. Appendix~\ref{sec:technical_details} contains technical details about our approach. \tomasz{check the tokenizers trained with Sentencepiece library. As done by Jirka.}


\section{Measuring Tokenizer Properties}
\label{sec:tokenization_measures}

This section presents our in-depth analytical approach to evaluate different aspects of multilingual tokenization. We introduce non-parametric measures that describe the key properties of multilingual tokenizers: quality of vocabulary representation for particular languages and lexical overlap across languages.


We base our analysis on the empirical probability distribution of vocabulary units $v \in \mathcal{V}$ computed on training corpus for each language $l$:

\begin{equation}
    d_{l,\mathcal{V}}(v) = \frac{f(v,C_l)}{\sum_{v \in \mathcal{V}}f(v, C_l)}
\end{equation}
\label{eqn:vocab_distribution}

Function $f(v, C_l)$ is the number of occurrences of a vocabulary unit $v$ in monolingual training corpus $C_l$.

\subsection{Vocabulary Allocation}

We aim to quantify how well multilingual vocabulary represents meaningful lexical units of particular languages. Our intuition is that a good lexical representation is obtained when: 1. It uses a vast portion of multilingual vocabulary, and thus a larger part of 
 the embedding layer is devoted to the language; 2. The text in the language is split into longer and potentially more meaningful tokens.

\paragraph{Vocabulary Allocation: Average Rank}

To measure the number of vocabulary units available for modeling specific languages, we propose an estimation of the average rank of vocabulary units in distribution over a monolingual corpus.\footnote{In this context, rank is the position of unit $v$ in the vocabulary $\mathcal{V}$ sorted in descending order by the probability distribution $d_{l,\mathcal{V}}$} This measure denotes how many tokens are typically considered by a language model that has access to language identity information but no context (probabilistic unigram LM).

 \begin{equation}
    \mathrm{AR}_{l,\mathcal{V}} = \sum_{v \in \mathcal{V}} \rank(v, d_{l,\mathcal{V}}) d_{l,\mathcal{V}}(v)
\label{eqn:avg-rank}
\end{equation}

Our intuition is that model will have better information about the language's lexicon when vocabulary is distributed over a larger number of tokens as more parameters of the input embedding layer would be allocated to represent language-specific features. Moreover, larger vocabularies tend to cover longer and more meaningful units.


\paragraph{Vocabulary Allocation: Characters per Token}

In line with previous intuition, longer tokens have a more meaningful representation. Therefore, we measure text fragmentation by computing the average number of characters for a vocabulary unit in monolingual corpus $C_l$.:

\begin{equation}
    \mathrm{CPT}_{l,\mathcal{V}}  = \frac{|C_l|}{|T_{\mathcal{V}}(C_l)|}
\end{equation}

$T_{\mathcal{V}}(C_l)$ is the tokenization of the corpus with vocabulary $\mathcal{V}$; $|C_l|$ is the size of the corpus measured as the number of characters. 
We choose the number of characters as the unit to relate to because it's not susceptible to cross-lingual differences regarding word boundaries and the average length of words. Still, the amount of information conveyed by a single character varies largely with the writing systems, e.g., texts written in logographic scripts (e.g., Chinese, Japanese) tend to be shorter in the number of letters than similarly informative ones in the phonetical script (e.g., Latin) \cite{perfetti_orthography_2005}.

\subsection{Vocabulary Overlap}

Another important property of multilingual vocabulary is sharing lexical units across languages. Previous works claimed that vocabulary overlap improves cross-lingual transfer for learning downstream tasks \cite{pires_how_2019, wu_beto_2019}. We measure overlap as the divergence between corpora distributions $d_l$ (defined in equation~\ref{eqn:vocab_distribution}).  
We use the Jensen-Shanon divergence.\footnote{In  NLP literature, JSD is also known as ``information radius'' \cite{manning_foundations_2001}.} We apply JSD because it is symmetric and applicable for distribution with different supports. The latter is often the case when distributions are estimated for languages with distinct writing systems.

\begin{multline}
    \mathrm{JSD}(d_{l1,\mathcal{V}} || d_{l2, \mathcal{V}})  = \\
     = \frac{1}{2} \sum_{v \in \mathcal{V}} d_{l1,\mathcal{V}}(v) \log_2{\frac{d_{l1,\mathcal{V}}(v) }{m_{l1,l2,\mathcal{V}}(v)}} + \\
    + \frac{1}{2} \sum_{v \in \mathcal{V}} d_{l2,\mathcal{V}}(v) \log_2{\frac{d_{l2,\mathcal{V}}(v)} {m_{l1,l2,\mathcal{V}}(v)}} 
\end{multline}

where:

\begin{equation}
    m_{l1,l2,\mathcal{V}} =  \frac{1}{2} d_{l1,\mathcal{V}} + \frac{1}{2} d_{l2,\mathcal{V}}
\end{equation}

JSD is bounded in the range $0$ to $1$. The lower the value, the larger the overlap across corpora.

Another possibility to quantify overlap is to count unique vocabulary units appearing in tokenized texts across languages.
The advantage of divergence is that it reflects the frequency of shared tokens across corpora.
It is also less affected by the choice of the data size used for estimating empirical probability distributions ($d_l$).






\section{Evaluating Language Modeling and Downstream Tasks}
\label{sec:lm}

 In this section, we present the tasks and measures for evaluation of multilingual language models trained with different tokenizers.
 

\subsection{Language Modeling}

We evaluate the masked language modeling performance with mean reciprocal rank:


\begin{multline}
 \mathrm{MRR} = \frac{1}{N}\sum_{i=1}^N  \frac{1}{\rank(x_i,\hat{P}(\cdot|X\setminus x_i))}
\end{multline}

where $\hat{P}(\cdot|X\setminus x_i)$ is the probability over vocabulary of predicting token $x_i$ by the model given its context: $X\setminus x_i$.

\subsection{Downstream Evaluation}

 The downstream tasks are taken from the XTREME \cite{hu_xtreme_2020}, which is the collection of diverse datasets with predefined splits used to evaluate multilingual models' representation.

 We probe the models' output representation to evaluate how useful the learned representation is for the downstream tasks.
 Only an additional linear layer is trained for the task, while the base model representation is frozen. The approach is suitable for evaluating how well the pre-trained model encodes linguistic phenomena as it does not change parameters learned in pre-training in contrast to regular fine-tuning \cite{conneau_what_2018, belinkov_probing_2021}.

\paragraph{Word-level Tasks}
The first set of tasks covers classification on a single word or word pair level. The probe is a linear layer taking word representations on input and outputting one of the classes. For word representations, we take the model's output embedding of the first subwords. 
We evaluate the results with an F1 score averaged across classes (macro-average).

We test syntactic tasks: \textbf{Part of Speech} and \textbf{Dependency labeling} on Universal Dependencies \cite{nivre_universal_2017} and \textbf{Named Entity Recognition} on Wikiann dataset \cite{pan_cross-lingual_2017}. In dependency labeling, we use edge probe \cite{tenney_what_2019} on top of the representation of two words connected by the dependency arc.

\paragraph{Sentence-level Tasks}

In this set of tasks, we examine whether the model learns sentence-level representations that capture its semantics and can be transferred across languages. To obtain this sentence embedding, we average the model's output representation across all the tokens in the sentence. 

We evaluate \textbf{Natural Language Inference} on XNLI dataset \cite{conneau_xnli_2018} and \textbf{Sentence Retrieval} on Tatoeba bitext corpus  \cite{artetxe_massively_2019}. For NLI, we use edge probing. Sentence retrieval is solved by an unsupervised algorithm matching sentences based on their cosine similarity.
In Appendix~\ref{sec:app_downstream}, we provide details of the datasets and probe training.

\subsubsection{In-language vs. Cross-lingual Transfer}

For all the downstream tasks, except sentence retrieval, we compute in-language performance by training the probe and evaluating it on held-out test data in the same language. We quantify cross-lingual transfer by training a probe on one language (source) and evaluating it on the test set for another language (target). 

\section{Experiments and Results}
\label{sec:experiments}
\pdfoutput=1

\begin{table}
\centering
\scriptsize
\begin{tabular}{llcccccc}
\toprule
     &        &                ar &                tr &                zh &                el &                es &                en  \\
\midrule
\multirow{4}{*}{AR} & Unigram &              2129 &              2719 &              \bf{5919} &              2070 &              1439 &              1513  \\
     & BPE &              2972 &              3226 &              4294 &              \bf{2907} &              \bf{2220} &              \bf{2143}  \\
     & NoOverlap &              2537 &              2653 &              2090 &              2065 &              1661 &              1597 \\
     & TokMix &              \bf{3485} &              \bf{4167} &              3961 &              2639 &              1999 &              1898  \\
\midrule
\multirow{4}{*}{CPT} & Unigram &              3.16 &              4.01 &              1.84 &               3.5 &              3.88 &              3.91 \\
     & BPE &              \bf{3.7} &              4.19 &              \bf{2.03} &              \bf{3.97} &              \bf{4.34} &              \bf{4.22}  \\
     & NoOverlap &              3.53 &              4.19 &              1.56 &              3.81 &              4.15 &              4.15 \\
     & TokMix &               \bf{3.7} &              \bf{4.45} &              1.73 &               3.9 &              4.24 &              4.18  \\ 
     \bottomrule
\end{tabular}
\caption{Values of \va~measures for 4 tokenizers trained on the small language set. The highest values for each language are bolded.}
\label{tab:vocab_allocation}
\end{table}
\pdfoutput=1
\begin{figure}[tb!]
    \centering
    \includegraphics[width=\linewidth]{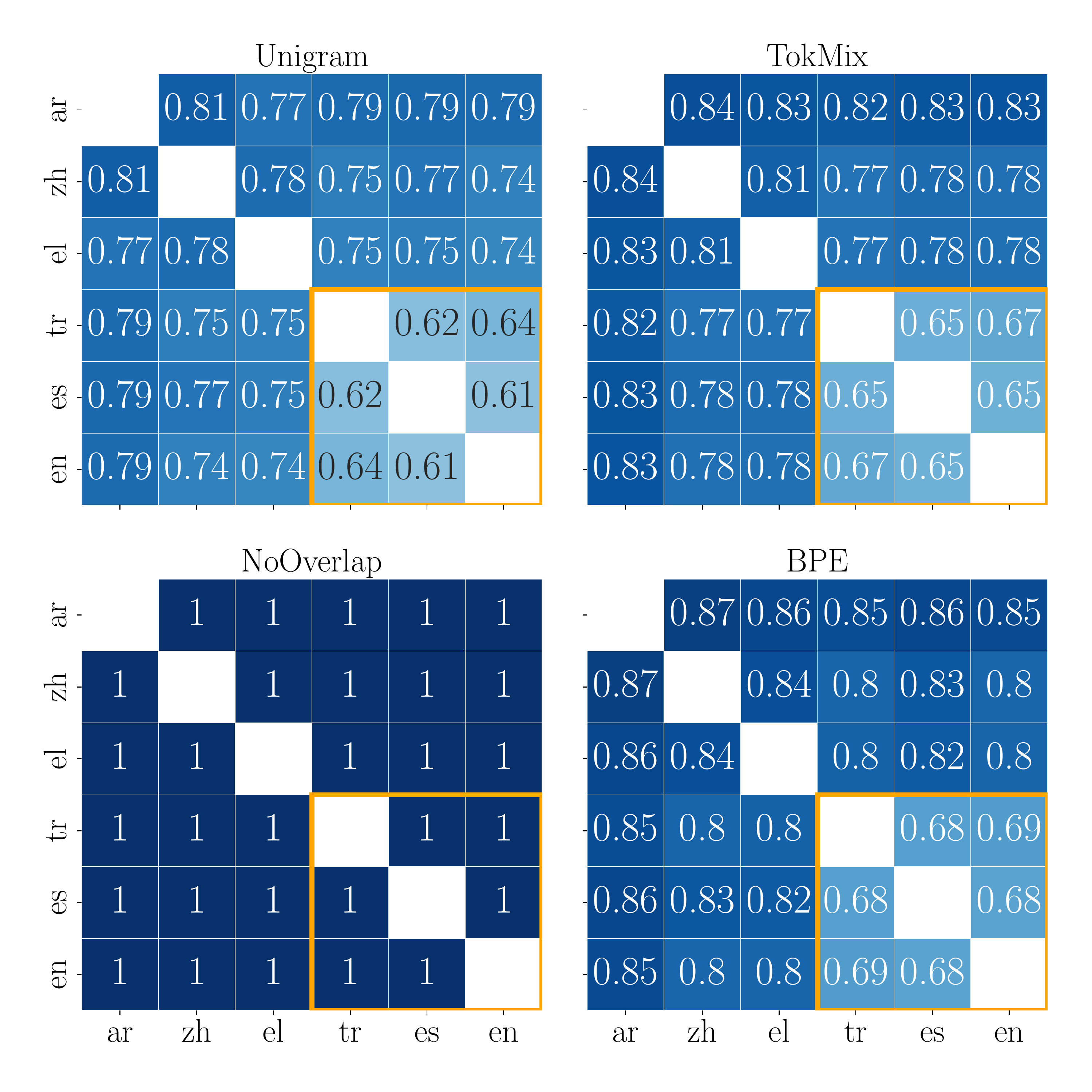}
    
    \caption{\emph{Vocabulary overlap} measure: Jensen-Shanon divergence for four tokenization methods. Orange square in the bottom right groups the languages with the same script (Latin).}
    \label{fig:jsd_6l}
\end{figure}

We train four tokenizers for the smaller set of diverse 6 languages (en, es, tr, el, zh, ar) using existing methods: Unigram, BPE, and our methods for monolingual tokenizer merging: \textsc{NoOverlap}, \textsc{TokMix}. Using these tokenizers, we then train four models\footnote{Details about the pretraining and probing procedures are described in Appendix~\ref{sec:model_architecture}} following the settings of XLM-Roberta \cite{conneau_unsupervised_2019} which we then use for the probing experiments.

In Section~\ref{sec:tokenizer_results}, we analyze the distribution of learned vocabulary units and compute \va~and \vo~measures described in Section~\ref{sec:tokenization_measures}.
Then in Section~\ref{sec:lm_results}, we evaluate the models' performance measures introduced in Section~\ref{sec:lm} and compare them with the measures for tokenizers.

Subsequently, we repeat the analysis for the broader set of 20 diverse languages (including six mentioned earlier and: he, ka, ur, hi, mr, th, ta, te, bg, ru, sw, vi, fr, de) with three tokenization methods used in three pre-trained models. In this setting, we do not use \textsc{NoOverlap} tokenizer, which cannot be trained effectively due to the necessity of constraining vocabulary for each language to $\frac{N}{L}=6,000$.

\subsection{Evaluation of Tokenizers' Properties}
\label{sec:tokenizer_results}

\begin{table*}[!htb]
\centering
\begin{subtable}{1\linewidth}
  \centering
  \footnotesize
    \begin{tabular}{lccccccc}
    \toprule
    & \multicolumn{2}{c}{\bf{V. Allocation}}  &  \bf{MLM} &               \bf{NER} &               \bf{POS} &                \bf{Dep. labeling} &              \bf{NLI} \\
     & (AR) &   (CPT) &   (MRR) &               (F1) &               (F1) &                (F1) &              (Acc) \\
    \midrule
    Unigram   &      2042 &  3.17 &  42.0 &  62.8 $_{\pm0.1}$ &  57.1 $_{\pm0.2}$ &  48.1 $_{\pm0.4}$ &  \bf{\underline{53.4}} $_{\pm0.5}$ \\
    BPE       &      2193 &  \bf{\underline{4.47}} &  35.6 &  \bf{\underline{70.4}} $_{\pm0.1}$ &  \bf{68.9} $_{\pm0.2}$ &  \bf{58.7} $_{\pm0.4}$ &  \bf{53.3} $_{\pm0.3}$ \\
    NoOverlap &      1829 &  3.16 &  \bf{\underline{42.7}} &  69.4 $_{\pm0.1}$ &  \bf{\underline{69.2}} $_{\pm0.2}$ &  \bf{\underline{58.8}} $_{\pm0.3}$ &  \bf{53.0} $_{\pm0.4}$ \\
    TokMix    &      \bf{\underline{2198}} &  3.34 &  38.7 &  \bf{70.2} $_{\pm0.1}$ &  67.3 $_{\pm0.1}$ &  57.3 $_{\pm0.4}$ &  \bf{53.3} $_{\pm0.4}$ \\
    \bottomrule
    \end{tabular}
    \caption{6 languages}
    \label{tab:in_lang_avg}
\end{subtable} \\
\begin{subtable}{1\linewidth}
  \centering
    \small
    \begin{tabular}{lccccccc}
    \toprule
    & \multicolumn{2}{c}{\bf{V. Allocation}}  &  \bf{MLM} &               \bf{NER} &               \bf{POS} &                \bf{Dep. labeling} &              \bf{NLI} \\
     & (AR) &   (CPT) &   (MRR) &               (F1) &               (F1) &                (F1) &              (Acc) \\
    \midrule
    Unigram   &       623 &  2.89 &  \bf{\underline{52.6}} &  58.9 $_{\pm0.2}$ &  54.0 $_{\pm0.4}$ &  43.7 $_{\pm0.4}$ &  \bf{53.2} $_{\pm0.3}$ \\
    BPE       &       \bf{\underline{809}} &  \bf{\underline{3.43}} &  40.5 &  \bf{\underline{66.3}} $_{\pm0.2}$ &  \bf{\underline{67.3}} $_{\pm0.4}$ &  \bf{\underline{54.5}} $_{\pm0.5}$ &  \bf{\underline{53.5}} $_{\pm0.3}$ \\
    TokMix    &       689 &  3.23 &  44.8 &  65.4 $_{\pm0.3}$ &  \bf{66.5} $_{\pm0.4}$ &  \bf{53.9} $_{\pm0.5}$ &  \bf{52.3} $_{\pm0.3}$ \\
    \bottomrule
    \end{tabular}
    \caption{20 languages}
    \label{tab:in_lang_avg_20l}
\end{subtable} 
\caption{Avearged results of evaluation for in-language properties and tasks. Each probing result is an average of 5 random seeds (for 6 languages) and 3 random seeds (for 20 languages). The best value in each metric is underlined, and bolded results are closer than the sum of standard deviations from the optimal value.}
\end{table*}
\begin{table}
\centering

\begin{tabular}{lccc}
\toprule
 & \multicolumn{2}{c}{\bf{V. Allocation}} & \bf{MLM} \\
 & (AR) &  (CPT)  &  (MRR) \\
\midrule
CPT    &    \bf{0.790} &     - &  - \\
MRR    &  \bf{-0.723} &  \bf{-0.913} &  - \\
NER    &   \bf{0.394} &   \bf{0.657} &  \bf{-0.745} \\
POS    &     0.320 &   \bf{0.724} &  \bf{-0.754} \\
Dep l. &     0.266 &   \bf{0.675} &  \bf{-0.695} \\
NLI   &    \bf{0.56} &    0.388 &  \bf{-0.437} \\ 
\bottomrule
\end{tabular}
\caption{Spearman correlations between task coefficients for in-language results and tokenizer measures. Statistically significant correlations ($p<0.01$) are bolded. Computed for 20 languages.}
\label{tab:corr_in_lang_20l}
\end{table}

\paragraph{\emph{Vocabulary allocation} largely varies throughout languages and tokenization methods.}

Table \ref{tab:vocab_allocation} shows that the average rank noticeably differs  across languages. The highest AR is observed for Chinese, which is caused by the fact that logographic scripts require an extensive vocabulary capacity to encode all characters. 

Multilingual \va~is highly dependent on 
the tokenization method used. Vocabulary learned  with Unigram underperforms BPE and \textsc{TokMix} in both average rank and character per token. Table \ref{tab:in_lang} presented in the Appendix shows that this trend exists throughout languages except for Chinese. This suggests that our vanilla Unigram is a suboptimal multilingual vocabulary learner.

It is important to note that \textsc{NoOverlap} scores even lower than Unigram in the \va~measures due to the limited vocabulary size for each language and disallowing overlap. However, as shown in the next sections, LM trained with this tokenizer can achieve good results on some tasks.

\paragraph{The choice of tokenization method affects \vo{}.}

Figure~\ref{fig:jsd_6l} shows Jensen-Shanon divergencies between the vocabularies of six languages. We observe that the highest cross-lingual overlaps appear in the vocabulary obtained by Unigram, followed by \textsc{TokMix}, and BPE. Expectedly, we do not observe overlaps for \textsc{NoOverlap}'s setting ($\mathrm{JSD}=1$).

Jensen-Shanon divergence is a good predictor of whether the languages share the script. For all tokenization methods, the divergence is significantly smaller in the bottom-right square grouping of the languages using Latin script. This effect is even more visible in the visualization of JSD computed for twenty languages (Figure~\ref{fig:jsd_20l} in Appendix~\ref{sec:app_results}).

\subsection{Tokenizer Properties Impact Language Model's Performance}
\label{sec:lm_results}

\begin{table*}[!htb]
\centering
\footnotesize
\begin{subtable}[t]{.55\linewidth}
    \begin{tabular}[t]{llcccc}
        \toprule
                &        &       Different &       Same &               All \\
        Metric & Tokenizer &       script       &     script         &     transfers          \\
        \midrule
        \multirowcell{4}[0pt][l]{\textbf{Overlap} \\ (JSD)} & Unigram &              \bf{\underline{0.77}} &             \bf{\underline{0.62}} &             \bf{\underline{0.74}} \\
                & BPE &              0.83 &              0.68 &               0.8 \\
                & NoOverlap &               1.0 &               1.0 &               1.0 \\
                & TokMix &               0.8 &              0.65 &              0.77 \\
        \midrule
        \multirowcell{4}[0pt][l]{\textbf{NER} \\ (F1)} & Unigram &  31.3 $_{\pm0.4}$ &  55.4 $_{\pm0.2}$ &  36.1 $_{\pm0.4}$ \\
                & BPE &  \bf{\underline{33.5}} $_{\pm0.5}$ &  \bf{\underline{59.9}}$_{\pm0.2}$ &  \bf{\underline{38.7}} $_{\pm0.4}$ \\
                & NoOverlap &  32.0 $_{\pm0.5}$ &  48.6 $_{\pm0.4}$ &  35.3 $_{\pm0.5}$ \\
                & TokMix &  31.8 $_{\pm0.4}$ &  58.0 $_{\pm0.3}$ &  37.0 $_{\pm0.4}$ \\
        \cline{1-5}
        \multirowcell{4}[0pt][l]{\textbf{POS} \\ (F1)} & Unigram &  18.1 $_{\pm0.4}$ &  38.3 $_{\pm0.4}$ &  22.2 $_{\pm0.4}$ \\
                & BPE &  \bf{\underline{25.8}} $_{\pm0.5}$ &  \bf{40.8} $_{\pm0.4}$ &  \bf{\underline{28.8}} $_{\pm0.5}$ \\
                & NoOverlap &  20.1 $_{\pm0.5}$ &  \bf{\underline{41.9}} $_{\pm0.5}$ &  24.5 $_{\pm0.5}$ \\
                & TokMix &  21.9 $_{\pm0.4}$ &  40.4 $_{\pm0.3}$ &  25.6 $_{\pm0.4}$ \\
        \cline{1-5}
        \multirowcell{4}[0pt][l]{\textbf{Dep. labeling}\\ (F1)} & Unigram &  11.1 $_{\pm0.3}$ &  25.5 $_{\pm0.3}$ &  14.0 $_{\pm0.3}$ \\
                & BPE &  \bf{\underline{15.9}} $_{\pm0.4}$ &  \bf{27.0} $_{\pm0.4}$ &  \bf{\underline{18.1}} $_{\pm0.4}$ \\
                & NoOverlap &  12.8 $_{\pm0.4}$ &  \bf{\underline{27.8}} $_{\pm0.5}$ &  15.8 $_{\pm0.4}$ \\
                & TokMix &  12.6 $_{\pm0.5}$ &  26.1 $_{\pm0.3}$ &  15.3 $_{\pm0.5}$ \\
        \midrule
        \multirowcell{4}[0pt][l]{\textbf{NLI} \\ (Acc)} & Unigram &  \bf{42.2} $_{\pm0.7}$ &  \bf{43.7} $_{\pm0.7}$ &  \bf{42.5} $_{\pm0.7}$ \\
                & BPE &  \bf{\underline{42.4}} $_{\pm0.7}$ &  \bf{\underline{45.2}} $_{\pm0.8}$ &  \bf{\underline{43.0}}$_{\pm0.7}$ \\
                & NoOverlap &  37.3 $_{\pm0.6}$ &  37.1 $_{\pm0.5}$ &  37.2 $_{\pm0.6}$ \\
                & TokMix &  \bf{41.2} $_{\pm0.7}$ &  \bf{42.7} $_{\pm0.5}$ &  \bf{41.5} $_{\pm0.7}$ \\
        \cline{1-5}
        \multirowcell{4}[0pt][l]{\textbf{Retrieval} \\ (Acc)} & Unigram &              21.0 &              \bf{\underline{43.9}} &              25.6 \\
                & BPE &              20.9 &              40.7 &              24.9 \\
                & NoOverlap &              12.3 &              28.0 &              15.4 \\
                & TokMix &              \bf{\underline{23.0}} &              43.4 &              \bf{\underline{27.1}} \\
        \bottomrule
    \end{tabular}
    \caption{6 languages}
    \label{tab:X_lang}
\end{subtable}
\hspace{.02\linewidth}
\begin{subtable}[t]{.42\linewidth}
    \begin{tabular}[t]{lcccc}
    \toprule
               &       Different &       Same &               All \\
     Tokenizer &       script    &       script         &         transf       \\
    \midrule
    Unigram &              \bf{\underline{0.75}} &              \bf{\underline{0.58}} &              \bf{\underline{0.73}} \\
            BPE &              0.83 &              0.67 &              0.81 \\ \\
         TokMix &               0.8 &              0.64 &              0.78 \\
    \midrule
    Unigram &  33.2 $_{\pm0.5}$ &  50.7 $_{\pm0.6}$ &  35.4 $_{\pm0.5}$ \\
            BPE &  \bf{\underline{36.6}} $_{\pm0.6}$ &  \bf{\underline{54.3}} $_{\pm0.3}$ &  \bf{\underline{38.8}} $_{\pm0.5}$ \\ \\
            TokMix &  \bf{36.5} $_{\pm0.6}$ &  \bf{53.7} $_{\pm0.5}$ &  \bf{38.7} $_{\pm0.6}$ \\ 
    \cline{1-4}
    Unigram &  23.4 $_{\pm0.5}$ &  32.9 $_{\pm0.3}$ &  24.6 $_{\pm0.5}$ \\
            BPE &  \bf{\underline{30.5}} $_{\pm0.6}$ &  \bf{\underline{40.7}} $_{\pm0.4}$ &  \bf{\underline{31.8}} $_{\pm0.6}$ \\ \\
            TokMix &  \bf{29.2} $_{\pm0.5}$ &  \bf{40.4} $_{\pm0.3}$ &  \bf{30.7} $_{\pm0.5}$ \\ 
    \cline{1-4}
    Unigram &  13.0 $_{\pm0.6}$ &  15.6 $_{\pm0.5}$ &  13.4 $_{\pm0.6}$ \\
            BPE &  \bf{\underline{16.5}} $_{\pm0.6}$ &  \bf{19.2} $_{\pm0.5}$ &  \bf{\underline{16.9}} $_{\pm0.5}$ \\ \\
            TokMix &  \bf{16.0} $_{\pm0.5}$ &  \bf{\underline{19.4}} $_{\pm0.4}$ &  \bf{16.5} $_{\pm0.5}$ \\
    \midrule
    Unigram &  \bf{37.3} $_{\pm0.5}$ &  37.5 $_{\pm0.4}$ &  37.4 $_{\pm0.5}$ \\
            BPE &  36.2 $_{\pm0.5}$ &  \bf{38.7} $_{\pm0.5}$ &  \bf{36.7} $_{\pm0.5}$ \\ \\
            TokMix &  \bf{\underline{37.8}} $_{\pm0.5}$ &  \bf{\underline{39.2}} $_{\pm0.5}$ &  \bf{\underline{38.1}} $_{\pm0.5}$ \\
    \cline{1-4}
    Unigram &              \bf{\underline{44.1}} &              44.4 &              44.2 \\
            BPE &              \bf{\underline{44.1}} &              \bf{\underline{49.1}} &              \bf{\underline{45.1}} \\ \\
            TokMix &              42.8 &              46.9 &              43.6 \\
    \bottomrule
    \end{tabular}
    \caption{20 languages}
    \label{tab:X_lang_20L}
\end{subtable}
\caption{Averaged results of the evaluation for cross-language overlaps and transfers.  Each probing result is an average of 5 random seeds (for 6 languages) and 3 random seeds (for 20 languages). The best value in each metric is underlined, and bolded results are closer than the sum of standard deviations from the optimal value. }
\end{table*}

\paragraph{High \va{} improves downstream results for word-level tasks.}

In Table~\ref{tab:in_lang_avg}, we observe that the choice of the tokenization method significantly impacts the results for POS, dependency labeling, and NER. We presume it results from learning good lexical representations throughout languages, e.g., by BPE and \textsc{TokMix}. The higher \va{}~is especially beneficial for word-level tasks. Whereas the influence on the sentence-level task (NLI) is minimal.

Notably, the model instance with \textsc{NoOverlap} tokenizer achieves the best F1 in POS and dependency labeling despite underperforming in \va{}. It is the result of learning language-specific representation for tokens that is especially useful for syntactic tasks.

\paragraph{Better MLM performance doesn't bring improvement to downstream tasks.}

In Table~\ref{tab:in_lang_avg}, we observe that the models performing better on masked token prediction (MRR) tend to be worse on downstream tasks (POS and NER). It is the result of different average ranks. The higher it is, the more vocabulary units a language model needs to consider for masked token filling, making masked word prediction harder. At the same time, a high average rank means that the vocabulary is broader and contains lexical units important for downstream tasks. 

Again, this trend does not hold for the results for \textsc{NoOverlap} setting, in which the search space for the masked-word problem is limited to the language-specific tokens leading to the best performance in MLM and syntactic tasks (POS and dependency label prediction).

In Table~\ref{tab:corr_in_lang_20l}, we show that the strong relationship between \va{} (avg. rank and CPT) and LM performance (MRR) is statistically supported. The length of token units has a strong positive influence on POS, dependency labeling, and NER results ($r > 0.65$) and a negative influence on MRR ($r < -0.9$), while it does not significantly affect NLI results. The correlation between the average rank and MRR, NER scores is weaker but still significant. Moreover, it is significantly correlated with XNLI accuracy with a medium coefficient $r=0.56$, even though the changes in XNLI are low across tokenizers.

\paragraph{Impact of \vo~on cross-lingual transfer varies across tasks.}

\pdfoutput=1

\begin{figure*}[!htb]
    \centering
        
    \begin{subfigure}[b]{0.48\textwidth}
        \centering
        \includegraphics[width=\textwidth]{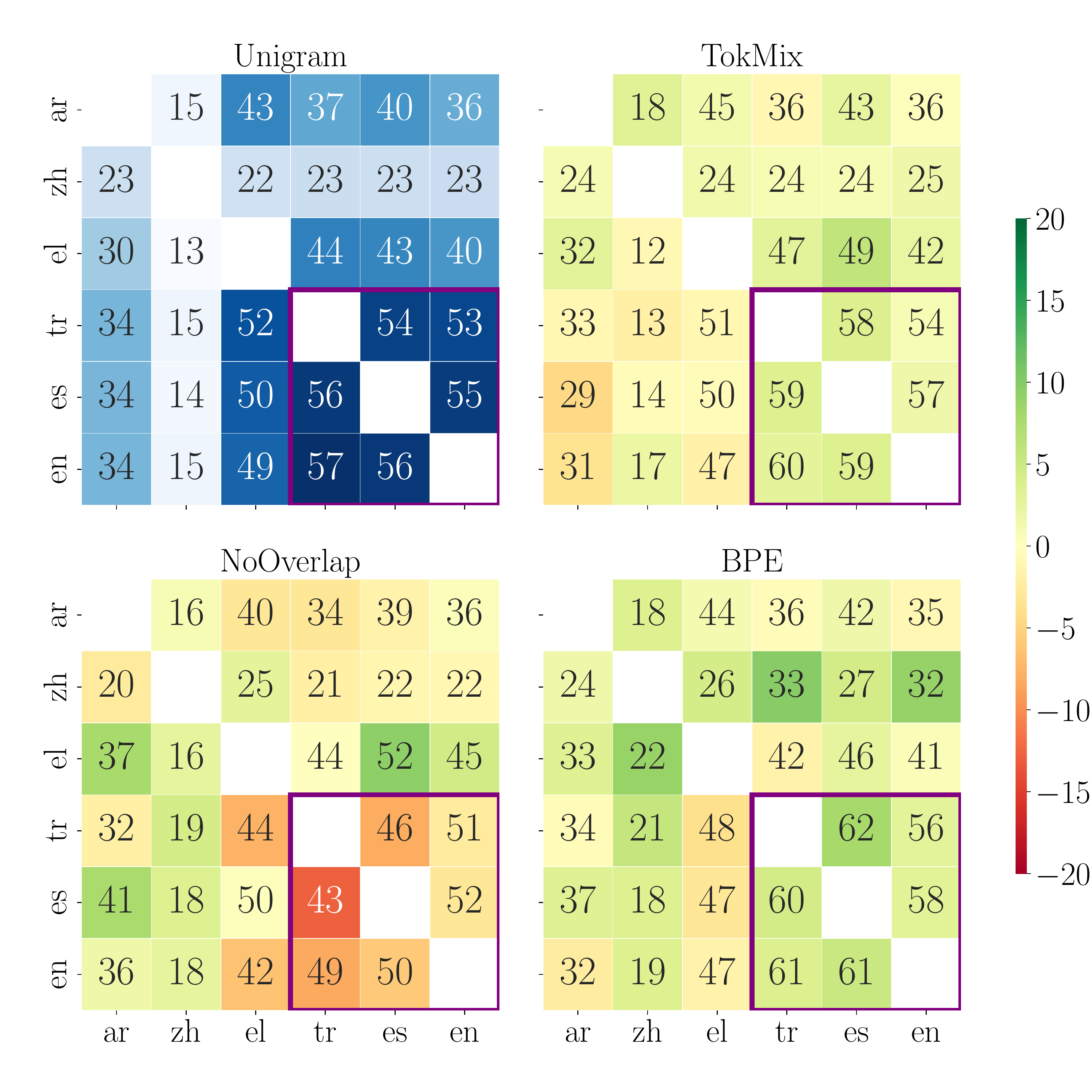}
        \caption{NER (F1)}
        \label{fig:ner_transfer}
    \end{subfigure}
    \hfill
    \begin{subfigure}[b]{0.48\textwidth}
        \centering
        \includegraphics[width=\textwidth]{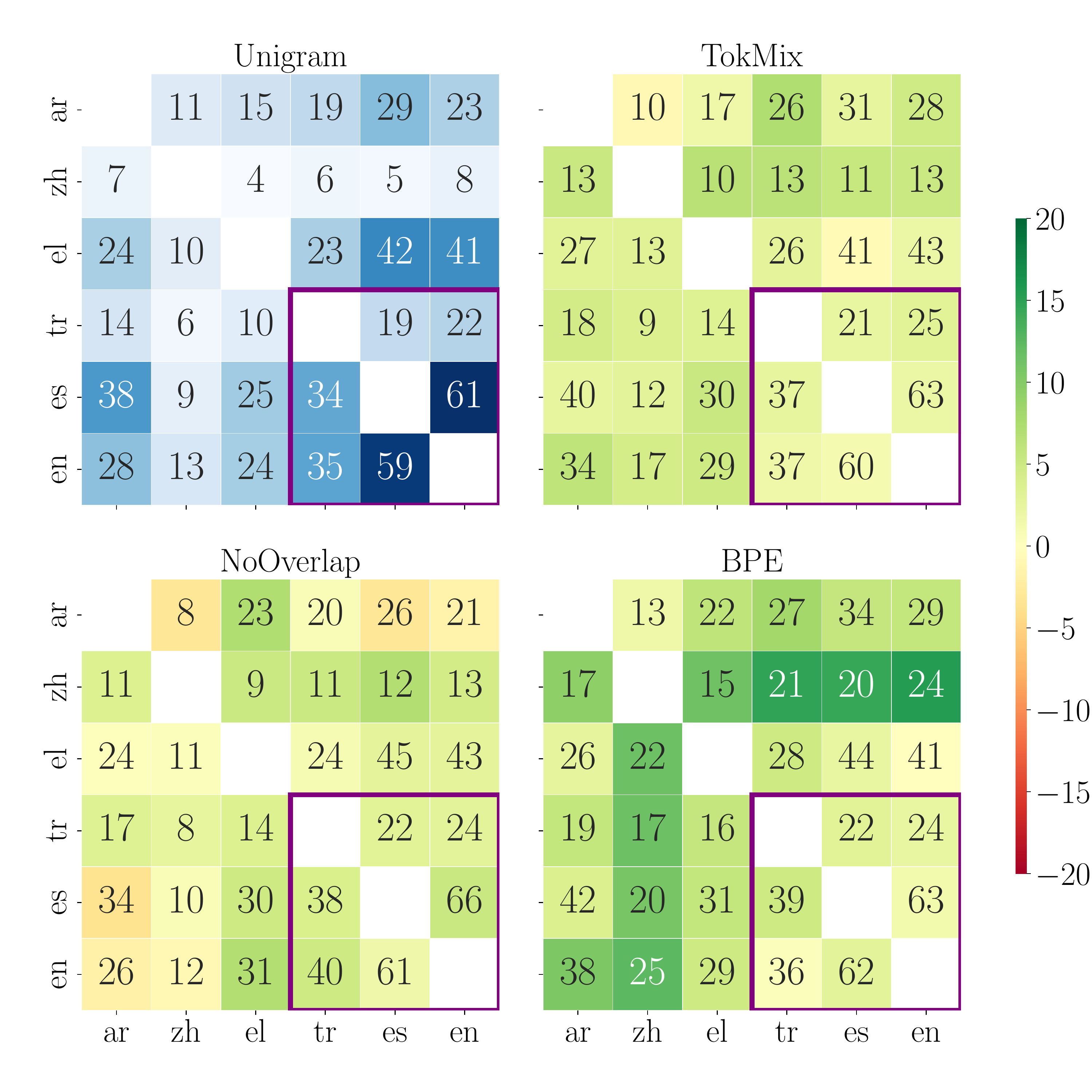}
        \caption{POS (F1)}
        \label{fig:pos_transfer}
    \end{subfigure}

    \caption{Cross-lingual transfer for POS and NER tasks. The absolute values are presented for the Unigram tokenizer. For other tokenization methods, the color scheme shows a difference from the Unigram algorithm. In the case of NER, we observe a drop in cross-lingual transfer for \textsc{NoOverlap} tokenization, especially for the same script pairs, suggesting that lexical overlap is an important aspect contributing to cross-lingual transfer for NER. We don't see similar drop in the case of Part of Speech tagging.}
\end{figure*}

\begin{table*}
\centering
\begin{tabular}{lcccccc}
\toprule
 & \bf{V. Overlap} & \multicolumn{2}{c}{\bf{V. Allocation SRC}}  & \multicolumn{2}{c}{\bf{V. Allocation TGT}} \\ 
 & (JSD) &  (AR)  &  (CPT) & (AR) & (CPT) \\ \midrule
NER     &    -0.111 &       \bf{0.249} &  \bf{ 0.33} &        0.209 &   \bf{0.28} \\
POS     &   \bf{0.395} &       \bf{0.365} &  \bf{0.547} &       \bf{0.489} &  \bf{0.653} \\
Dep l.  &   \bf{0.463} &         0.19 &  \bf{0.425} &       \bf{0.249} &   \bf{0.44} \\
NLI    &  \bf{-0.516} &       \bf{0.421} &   0.203 &       \bf{0.297} &    0.103 \\
Retrieval &  \bf{-0.648} &       \bf{0.235} &    0.082 &       \bf{0.238} &    0.085 \\
\bottomrule
\end{tabular}
\caption{Spearman correlations between cross-lingual transfer results and tokenization measures. \vo~is measured by JSD, we also measure the correlation with \va s of source and target language of the transfer directions. Statistically significant correlations ($p<0.01$) are bolded. Computed for six languages.}
\label{tab:corr_x_lang_6l}
\end{table*}

We observed that \textsc{NoOverlap} approach obtains competitive results for POS tagging 
. Surprisingly no vocabulary sharing also improves cross-lingual transfer in the task among languages with Latin script (shown in Table~\ref{tab:X_lang} and Figure~\ref{fig:pos_transfer}). We think that the reason behind the strength of \textsc{NoOverlap} approach is that some tokens have different meanings across languages, e.g.,  the word ``a'' is an indefinite article in English and a preposition in Spanish.

Nevertheless, vocabulary overlap is crucial to cross-lingual transfer in some tasks.  Especially NER within the same script languages (Figure~\ref{fig:ner_transfer}) and sentence-level tasks. For these tasks, \textsc{NoOverlap} significantly underperforms other tokenization methods. The drop within Latin script languages is in the range: $6.8$ - $11.3\%$ for NER and $12.7$ - $15.9\%$ for sentence retrieval. In these cases, usage of the same tokens can indicate that texts refer to the same entities across languages, e.g., names are usually the same strings in the languages sharing writing system. 
\newpage


Table~\ref{tab:corr_x_lang_6l} presents the correlations for cross-lingual transfer scores with JSD measuring \vo. The coefficient supports our previous observation that lower overlap (thus higher JSD) improves transfer for POS tagging and dependency labeling and deteriorates it for other tasks. Although, the correlation for NER is not significant. 
The \va s of source and target languages significantly influence the cross-lingual transfers. Similarly to the in-language correlations, the influence of 
character per token is more substantial on word-level tasks, while Average Rank affects sentence-level tasks to a larger extent. This observation underlines the importance of allocating a sufficient portion of vocabulary for low-resource for better cross-lingual transfer.
\footnote{We describe the correlation analysis in detail in Appendix~\ref{sec:correlation_analysis}.}

\paragraph{Results generalize to the larger set of languages.}

The key observation for six language sets holds in the model trained for twenty languages. Table~\ref{tab:in_lang_avg_20l} shows that BPE and \textsc{TokMix} obtain better \va~than Unigram leading to improved results for word-level downstream tasks (NER, POS, Dependency labeling). Due to the smaller vocab size to the language number ratio, average ranks decrease for all methods.

We observe in Table~\ref{tab:X_lang_20L} that the cross-language vocabulary overlap is the highest for Unigram and lowest for BPE, similar to the six languages settings. However, the association between \vo{} and the cross-lingual transfers is less pronounced. 





\section{Related Work}
\label{sec:related_work}

\paragraph{Importance of \vo{}.}
\citet{wu_beto_2019, pires_how_2019} claimed that multilingual overlap benefits cross-lingual transfer. In contrast to this work, they compare overlaps for different language pairs with only one tokenizer. We think that their observations may be confounded by the typological similarity between languages. In the following works, \citet{conneau_emerging_2020} found that sharing parameters in top layers is more important to multilingualism than same token embedding. Similar results were demonstrated by \citet{wang_multi-view_2021, dufter_identifying_2020} who show that in bilingual models, artificially removing \vo~(similarly to ours \textsc{NoOverlap}) does not deteriorate cross-lingual transfer. In contrast to many previous approaches, we used probing for evaluation because this method offers better insight into representation learned in pre-training. Similarly, our results, \citet{malkin_balanced_2022, limisiewicz_you_2022} observed that differences in scripts could, in some cases, improve the cross-lingual transfer in masked language modeling and for downstream tasks.

\paragraph{Importance of \va{}.}

The effect of \va{} on model performance was studied to a lower extent. \citet{zheng_allocating_2021} observed that limited vocabulary capacity allocated for specific languages impedes the downstream tasks' performance and thus proposed a method to obtain more balanced \va{} throughout languages. For the same purpose, \citet{chung_improving_2020} proposed a novel approach to generating multilingual vocabulary based on clustering the target languages and merging separate vocabularies. Recently, \citet{liang_xlm-v_2023} based on the elements of both approaches and increased vocabulary to train the XLM-V model, achieving better results than its predecessor (XLM-Roberta \citet{conneau_unsupervised_2019}).

In a monolingual setting, \citet{bostrom_byte_2020} argued that Unigram tokenization produces subword tokens that are more aligned with morphological units that bring improvement for downstream tasks. This contrasts with our finding of Unigram's underperformance when applied to a multilingual corpus.

\paragraph{Improving multilingual sub-word tokenization.}
\tomasz{Prthaps split this section into two subsequent ones}

\citet{patil_overlap-based_2022} proposed a modification to BPE algorithm that increases overlap between similar languages and benefits cross-lingual transfer. \citet{rust_how_2021} observed that models with dedicated monolingual tokenizers outperform multilingual ones. This observation can be utilized by adapting the embedding layer of the model for a target language \cite{pfeiffer_mad-x_2020, artetxe_cross-lingual_2020, minixhofer_wechsel_2022}. However, these approaches require language-specific modification of the model, limiting its multilingual aspect.

\paragraph{Alternatives to sub-word tokenization.}
 There are multiple  alternative approaches for inputting text into deep models, such as character-based representation \cite{clark_canine_2022}, byte input \cite{xue_byt5_2022}, or representing the input text as images \cite{salesky_robust_2021}. \citet{mielke_between_2021} summarize a wide range of methods and point out that they offer trade-offs and may be better suited for certain tasks or languages.

\section{Conclusions}
\label{sec:conclusion}
We introduced a new framework for the evaluation of multilingual subword tokenizers. We show that \va~is a crucial aspect affecting the results of many downstream tasks. Specifically, we have observed the following trends:
1. Including longer and more diverse vocabulary units (higher \va) improves in-language results and cross-lingual transfers for word-level tasks; 2. \vo~is beneficial for cross-lingual transfer in sentence-level tasks; 3. Among languages with the same script, \vo~improves transfer for NER and deteriorates it for POS and dependency labeling.
Our conclusions are in line with the observation of \citet{mielke_between_2021} that there is no ``silver bullet solution'' tokenizer suiting all purposes.

We release the code for measuring tokenizer properties: \url{github.com/tomlimi/entangled_in_scripts}. We believe that it will be a useful evaluation tool for the developers of models who can get a better insight into the tokenization method before computationally expensive model training.

\section*{Limitations}
\label{sec:limitations}

To achieve robust, unbiased results, we decided to train first on a smaller number of languages, fix our methodology and then confirm our findings on the full set of languages. This meant that two rounds of pretraining needed to be done and because of that, we scaled our models down for computational efficiency reasons.

Another limitation of our methodology is the choice to train linear probes on top of the contextualized word representations instead of the more common finetuning approach. Nevertheless, we think that probing gives better insight into the pre-trained model's representation.


\tomasz{Models are small.}
\tomasz{TokMix can lead to a larger portion of unknowns on training texts. Do a short overall analysis.}
\tomasz{The results may be less important on fine-tuning for languages with a high amount of data. However, it'll be still highly relevant for cross-lingual transfer.}
\tomasz{We do not use methods of tokenization regularization, such as distribution sampling in Unigram or BPE dropout.}

\section*{Ethics Statement}
We do not identify ethical risks connected to this work.

\section*{Acknowledgements}
We thank Jindřich Libovický, Martin Popel, Gabriel Stanovsky, and anonymous ACL reviewers for their valuable comments and suggestions for improvement. 
This work has been supported by grant 338521 of the Charles University Grant Agency.
We have been using language resources and tools developed, stored, and distributed by the LINDAT/CLARIAH-CZ project of the Ministry of Education, Youth and Sports of the Czech Republic (project LM2018101). 

\bibliography{eis}
\bibliographystyle{acl_natbib}

\clearpage
\appendix

\section{Technical Details}
\label{sec:technical_details}
\subsection{Tokenizer training details}

We use the Huggingface Tokenizers library for training the Unigram and BPE tokenizers. We kept the default values for the training parameters. Namely, for Unigram, we use a maximum piece length of 16 and a shrinking factor of 0.75. For BPE, we use alphabet size 1000 and minimum merge frequency 2. For all languages, we use SentencePiece \cite{kudo_sentencepiece_2018} for word segmentation techniques instead of language-specific word tokenizers. 

\subsection{Model Architecture and Pre-Training}
 \label{sec:model_architecture}

In this study, we employed the Huggingface library \cite{wolf_transformers_2020} to conduct all experiments. The model architecture is based on XLM-Roberta, although for our purposes, it was scaled down. Specifically, the size of the embeddings is 768, the number of attention layers is 8, and the number of attention heads is 6. The maximum sentence length is 128, and the vocabulary size is 120000. The number of parameters is 150M and, therefore, roughly 2 times smaller than the XLM-Roberta base model.

The model was pre-trained for 10 epochs with a batch size of 1024.
The learning rate was 5e-5 with linear decay and weight decay and 1\% warm-up steps. In pretraining, we used AdamW optimizer \cite{loshchilov_decoupled_2019}.

In total, we pretrained 7 models. The models were trained on 3 Nvidia GPUs. The probing experiments were run on 1 Nvidia GPU with 40GB of memory (Nvidia A40). The pretraining took about 17 hours for each 6-language model and 60 hours for the models trained on the full set of 20 languages.

We didn't pursue any extensive hyperparameter search efforts as this was not the focus of our work. We selected the best batch size and learning rates for the pre-training based on a few trials.

\subsection{Downstream Data and Training}
\label{sec:app_downstream}

The probes were for 30 epochs with early stopping and batch size 16. We used an initial learning rate of 2e-5. Other training parameters were the same as in pretraining. Probing experiments took between 5 to 180 minutes to complete on the same infrastructure as used for pretraining. We ran around 360 probe trainings.

\paragraph{POS}

We use Part of Speech annotations from Universal Dependencies \cite{nivre_universal_2017}. The dataset is available for 17 languages analyzed by us (not covered: Swahili, Thai, Georgian). Each word is assigned one of the 17 coarse POS tags.

\paragraph{NER}

We use Wikiann dataset \cite{pan_cross-lingual_2017} consisting of Wikipedias article with annotated named entities of three types: location, person, and organization in IOB2. Following XTREME, we use balanced data splits from \cite{rahimi_massively_2019}.

\paragraph{Dependency labeling}
As in Part of Speech, we use Universal Dependencies \cite{nivre_universal_2017} for the dependency relation annotations. We use the largest UD treebank available for each language.
For each word we predict one of the 37 universal relations to its head word. Because the relation is between two words, we use the concatenation of the two word representations along with their element-wise product as an input to the probe ($[h_{w1}; h_{w2}; h_{w1} \odot h_{w2}]$).

\paragraph{NLI}

We use XNLI dataset \cite{conneau_xnli_2018} for Natural Language Inference. We train the linear classification probe on top of the concatenation of two sentence vectors and their element-wise product: $[h_{s1}; h_{s2}; h_{s1} \odot h_{s2}]$. We predict one of two relations between the first of sentences (called premise): contradicts, entails, or is neutral to the second sentence (called a hypothesis). We evaluate XNLI with the accuracy of classification.

XNLI contains data for 15 languages (not covered: te, ta, mr, he, ka).

\paragraph{Sentence Retrieval}
We use up to 1,000 sentences aligned for pairs of languages from Tatoeba dataset \cite{artetxe_massively_2019}. For the pairs including English, we use the same sample as in XTREME data collection. For other pairs, we perform sampling ourselves. 

We compute the cosine similarity between sentence representations across languages and find the best alignment with the Hungarian algorithm\cite{kuhn_hungarian_1955}. We compute the accuracy as the number of correctly aligned sentences divided by the total number of sentences.

\section{In-depth Tokenizers Analysis}
\label{sec:app_tokenizers}
\pdfoutput=1
\begin{figure}[tb!]
    \centering
    \includegraphics[width=1.0\linewidth]{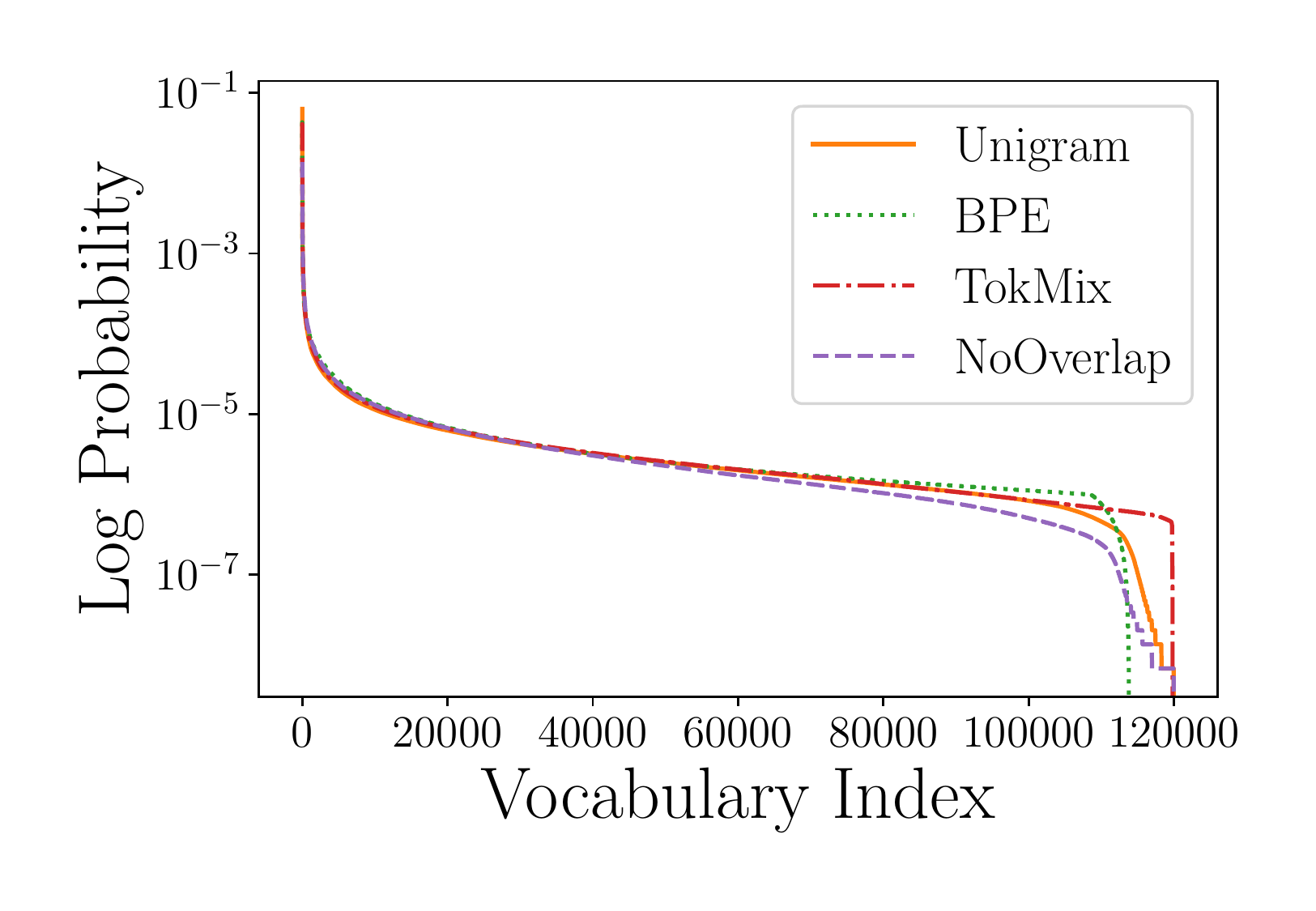}
    
    \caption{Log-probabilites of vocabulary units in decreasing order for four tokenization methods.}
    \label{fig:tokenizers_log_pdf}
\end{figure}

In Figure~\ref{fig:tokenizers_log_pdf}, we present the probabilities of vocabulary units, computed on concatenate six languages corpora, learned by different tokenization algorithms. Unigram and \textsc{NoOVerlap} use a bigger fraction of the vocabulary for rarely appearing tokens (with probability lower than $10^{-6}$). BPE and \textsc{TokMix} produce a vast set of tokens with probabilities in the range between $10^{-5}$ and $10^{-6}$. Interestingly, the former algorithm allocates about 6000 vocabulary entries to tokens not appearing in the corpora. 

\paragraph{BPE is better than Unigram in \va{} throughout languages.}

\pdfoutput=1
\begin{figure}[tb!]
    \centering
    \includegraphics[width=1.0\linewidth]{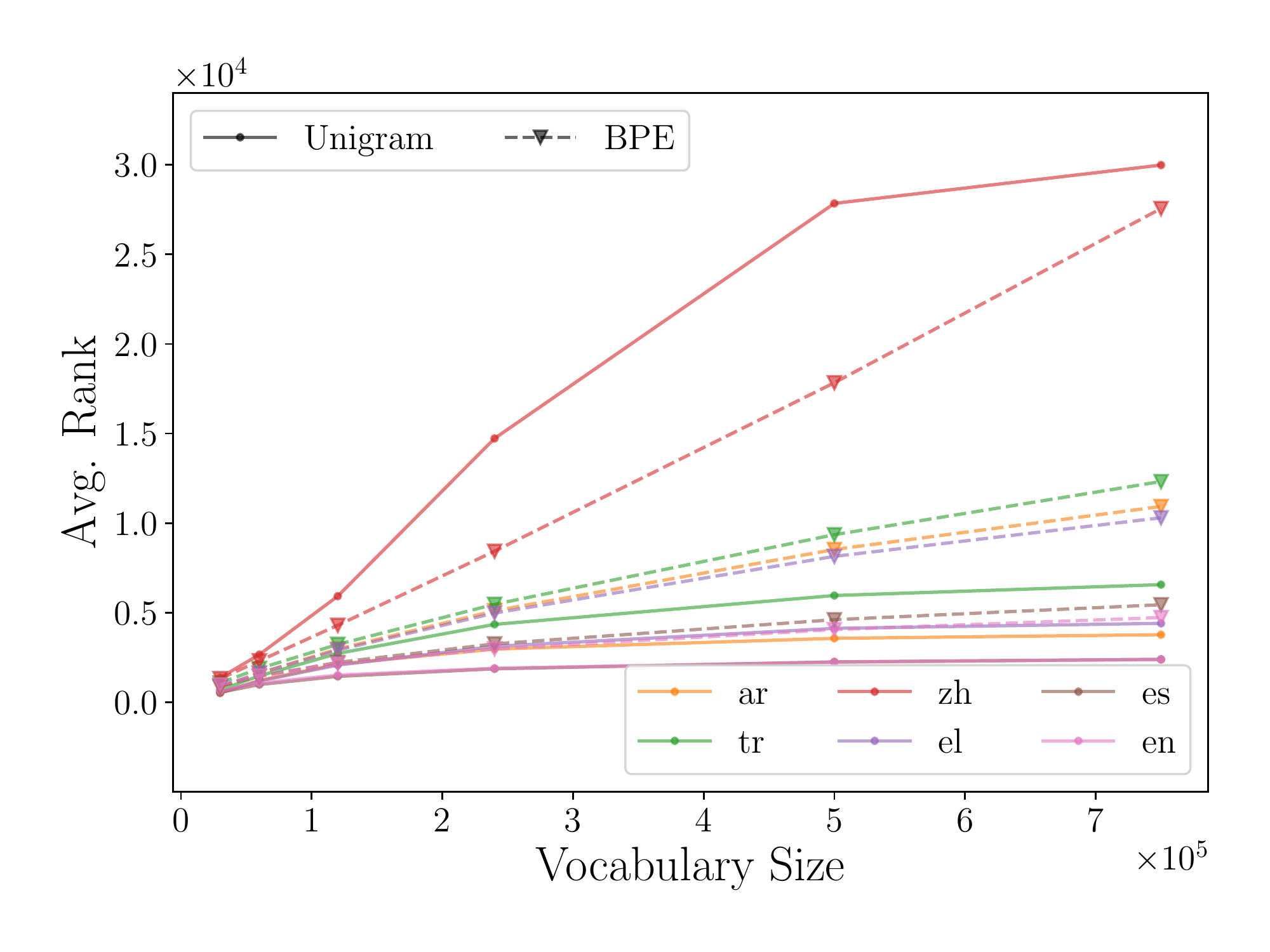}
    
    \caption{Avearage Rank measured for vocabularies of different sizes, obtained with BPE and Unigram algorithms.}
    \label{fig:ar_vocab}
\end{figure}
\pdfoutput=1
\begin{figure}[tb!]
    \centering
    \includegraphics[width=1.0\linewidth]{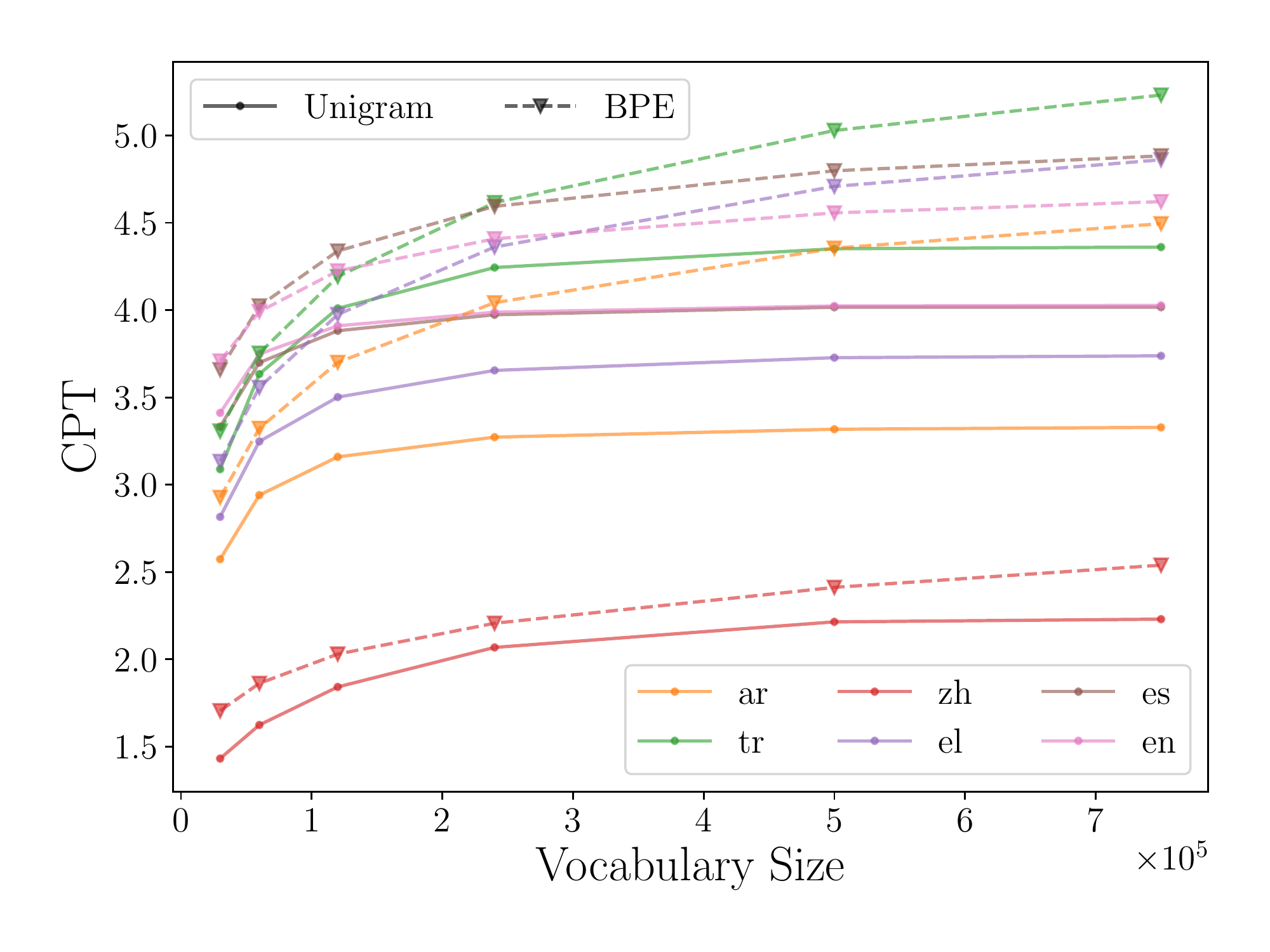}
    
    \caption{Characters per Token measured for vocabularies of different sizes, obtained with BPE and Unigram algorithms.}
    \label{fig:cpt_vocab}
\end{figure}

To support this claim, we train Unigram and BPE tokenizers for different vocabulary sizes. We observe that both the average rank (Figure~\ref{fig:ar_vocab}) and CPT (Figure~\ref{fig:cpt_vocab}) stop rising for vocab sizes above 250,000 (except for Chinese). For BPE, the metrics still steadily rise after this threshold, which makes it overperform Unigram for most languages.

We think that the reason why Unigram does not learn valuable tokens after this point is the way the initial vocabulary is constructed, i.e., it is the set of all character n-grams appearing in the corpus with n lower than 16. In contrast to BPR, Unigram's vocabulary won't cover longer words than 16 characters, which are useful in modeling some languages.

We believe that further work on identifying optimal strategies for multilingual tokenization is needed.

\paragraph{Vocabulary units preferred by tokenizers.}
\begin{table}[!tb]
\tiny
\begin{tabular}{@{}llll@{}}
\toprule
 &
  English &
  Turkish &
  Greek \\ \midrule
Unigram &
  \begin{tabular}[c]{@{}l@{}}s, ing, ed, \\ ly, d, If\end{tabular} &
  \begin{tabular}[c]{@{}l@{}}n, a, e, \\ k, s, i\end{tabular} &
  \begin{tabular}[c]{@{}l@{}}\textgreek{η, ς, ο},\\ \textgreek{α, ή, ει}\end{tabular} \\ \hline
BPE &
  \begin{tabular}[c]{@{}l@{}}▁the, ▁to, ▁of, \\ ▁and, ▁If,▁a\end{tabular} &
  \begin{tabular}[c]{@{}l@{}}▁o, ▁veyaim, im\\ inin, ası, esi\end{tabular} &
  \begin{tabular}[c]{@{}l@{}}\textgreek{▁η, ▁ο, ▁και,} \\ \textgreek{▁ή, ▁να, ▁στον}\end{tabular} \\ \bottomrule
\end{tabular}
\caption{List of units from Unigram and BPE vocabulary with the highest difference in frequency between tokenizers. The first row shows the tokens that appear more frequently in the corpus tokenized by Unigram and the second by the BPE tokenizer. We excluded punctuation marks and special characters from the list.}
\label{tab:learned_vocab_comparison}
\end{table}

In Table~\ref{tab:learned_vocab_comparison}, we show the tokens with the highest differences in empirical probabilities obtained with BPE and Unigram tokenizers for three languages. We see that Unigram prefers suffixes to prefixes. Also, it splits text more often into single, possibly due to lower \va.

\section{Supplementary Results}
\label{sec:app_results}

 \subsection{Visualizations}

\pdfoutput=1
\begin{figure}[tb!]
    \centering
    \begin{subfigure}[b]{1.0\linewidth}
        \centering
        \includegraphics[width=0.9\linewidth]{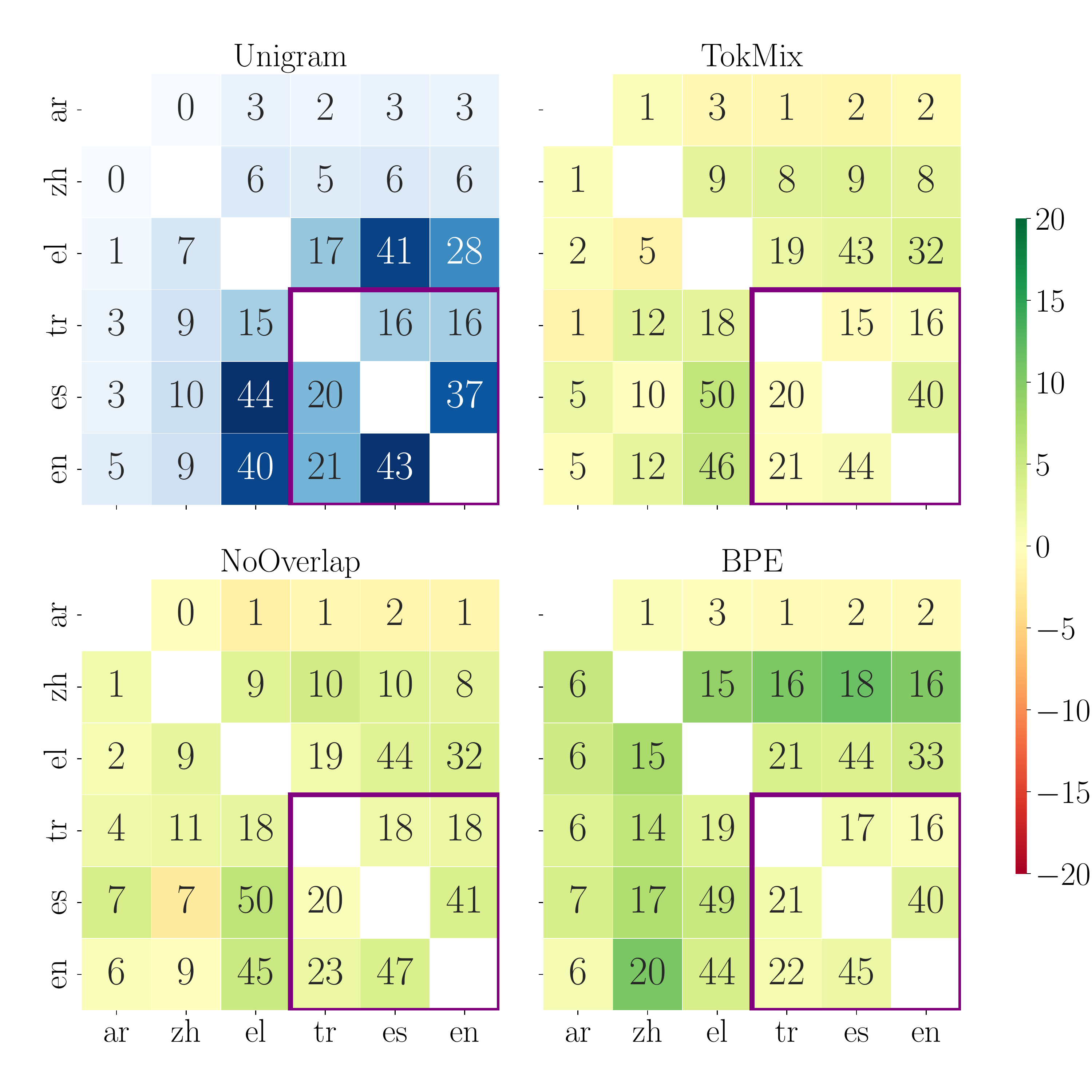}

        \caption{Dependency labeling}
        \label{fig:ud_transfer}
    \end{subfigure}
    \begin{subfigure}[b]{1.0\linewidth}
        \centering
        \includegraphics[width=0.9\linewidth]{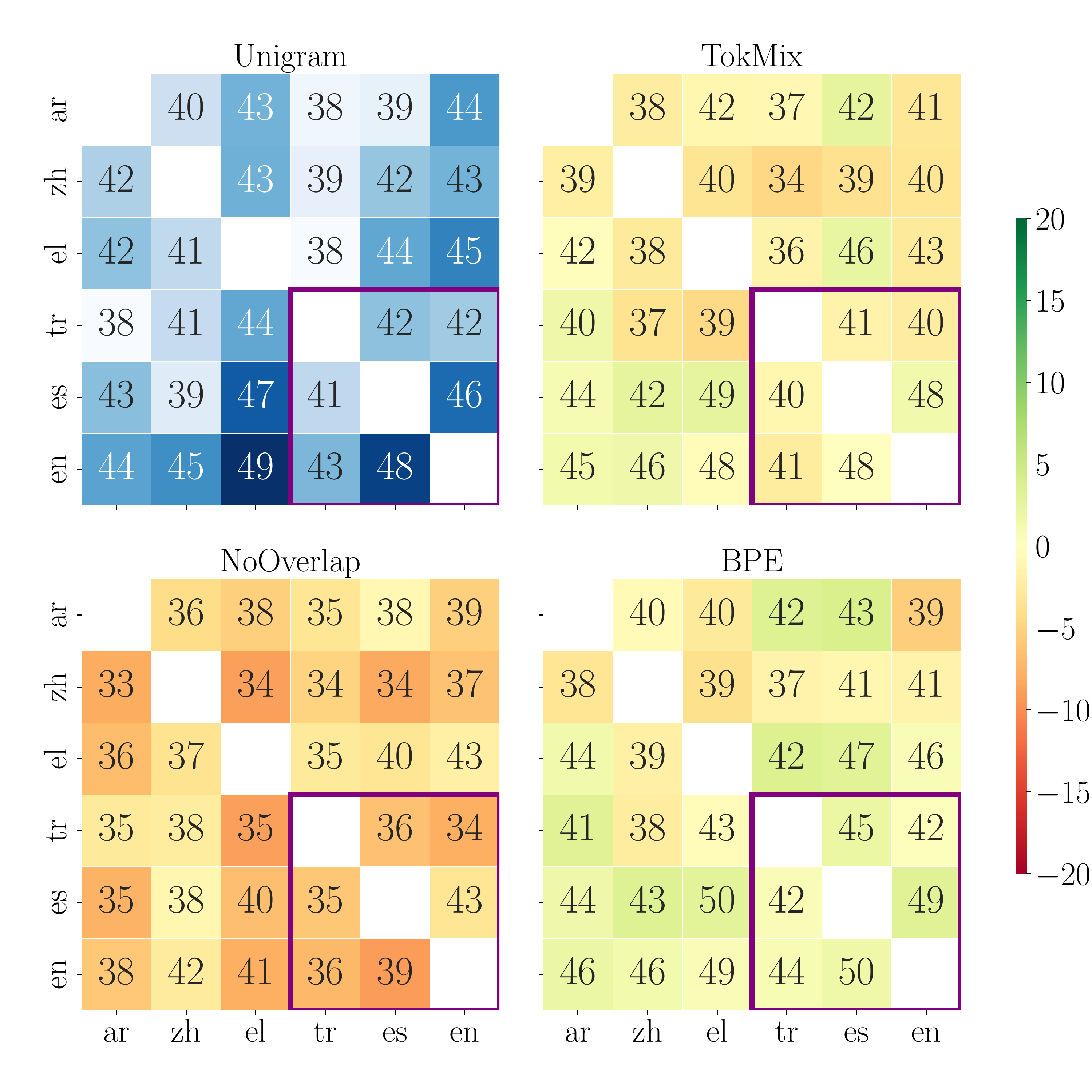}
        
        \caption{Natural Language Inference}
        \label{fig:xnli_transfer}
    \end{subfigure}
    \begin{subfigure}[b]{1.0\linewidth}
        \centering
        \includegraphics[width=0.9\linewidth]{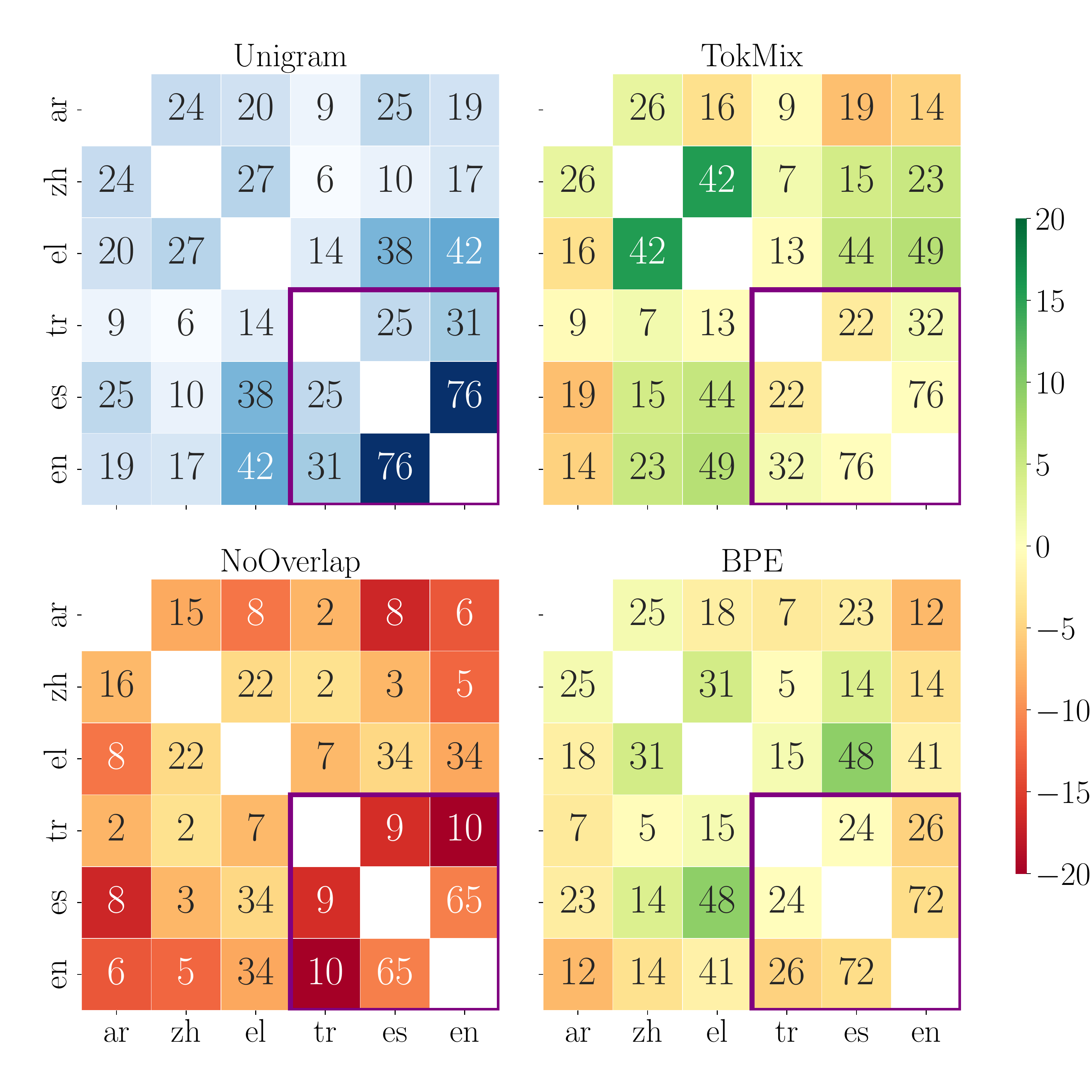}
        
        \caption{Sentence retrieval}
        \label{fig:tatoeba_transfer}
    \end{subfigure}
    \caption{The rest of the 6-language cross-lingual transfer results. The absolute values are presented for the Unigram tokenizer. For other tokenization methods, we show  the difference from the unigram algorithm.}
\end{figure}

 We present the additional visualization for 
 the results for transfers across six languages for the tasks not presented in the main text: Dependency labeling~\ref{fig:ud_transfer} and NLI cross-lingual accuracy~\ref{fig:xnli_transfer}, Sentence retrieval accuracy~\ref{fig:tatoeba_transfer}.

  \pdfoutput=1
\begin{figure*}[tb!]
    \centering
    \includegraphics[width=1.\linewidth]{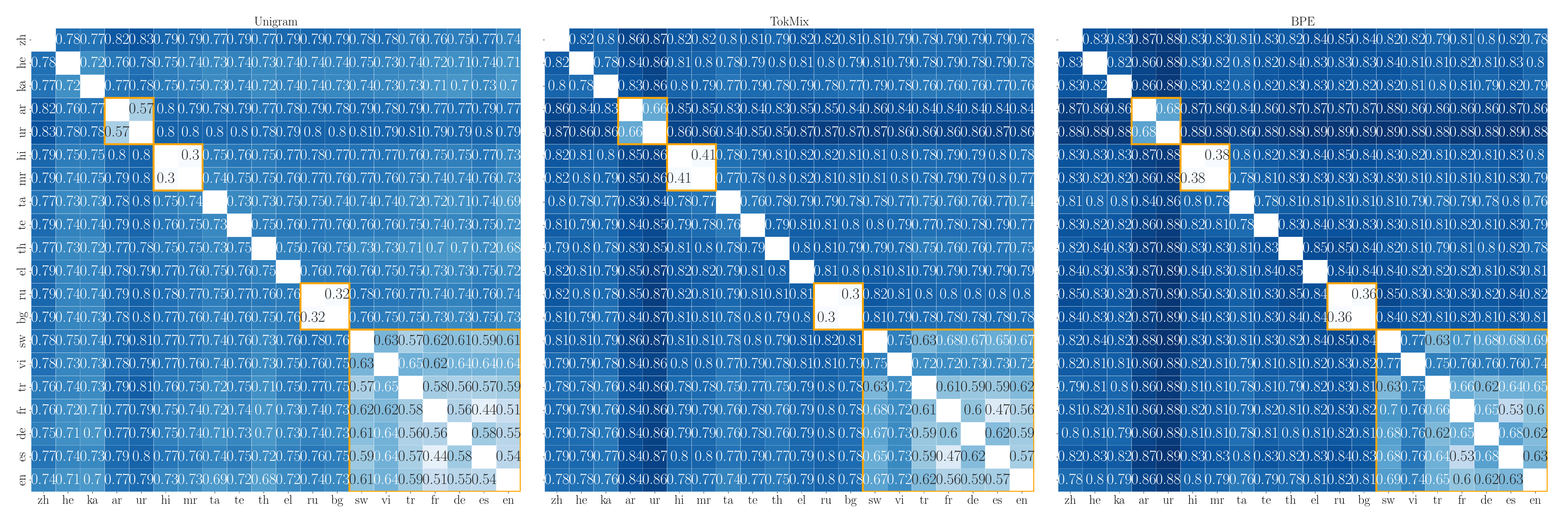}

    \caption{Jensen-Shanon divergence for three tokenization methods, computed on 20 languages.}
    \label{fig:jsd_20l}
\end{figure*}
    \pdfoutput=1
\begin{figure*}[tb!]
    \centering

    \begin{subfigure}[b]{1.0\linewidth}
        \centering
        \includegraphics[width=1.\linewidth]{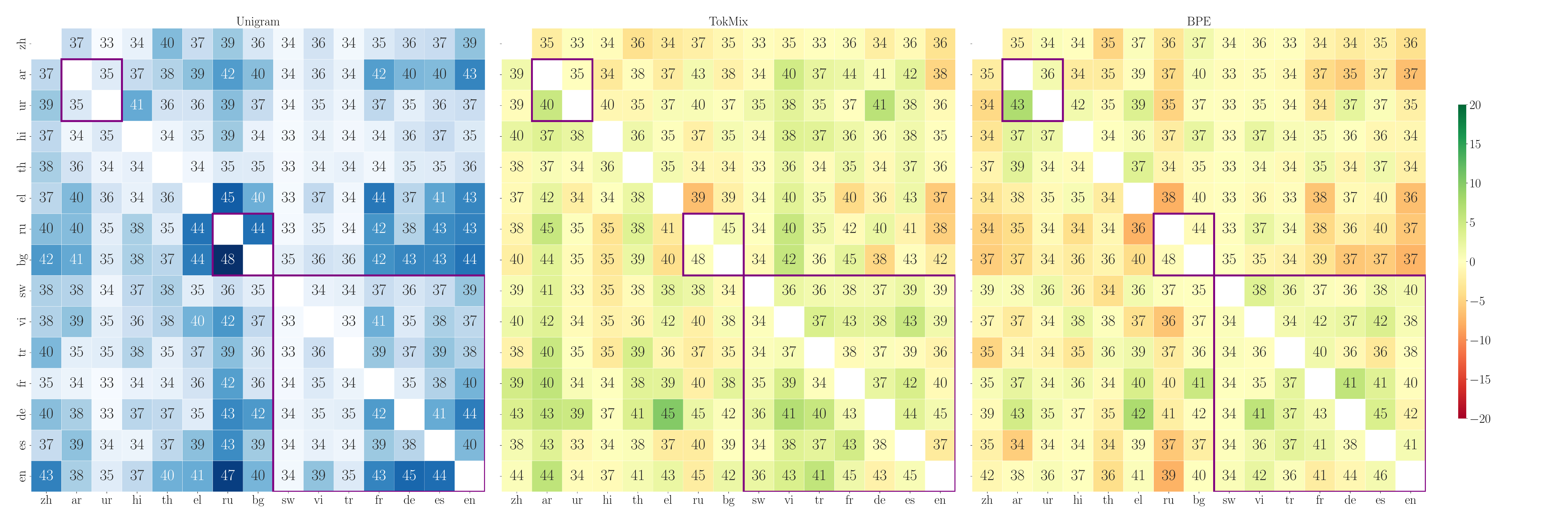}
    
        \caption{Natural Language Inference}
        \label{fig:xnli_transfer_20l}
    \end{subfigure}
    \begin{subfigure}[b]{1.0\linewidth}
        \centering
        \includegraphics[width=1.\linewidth]{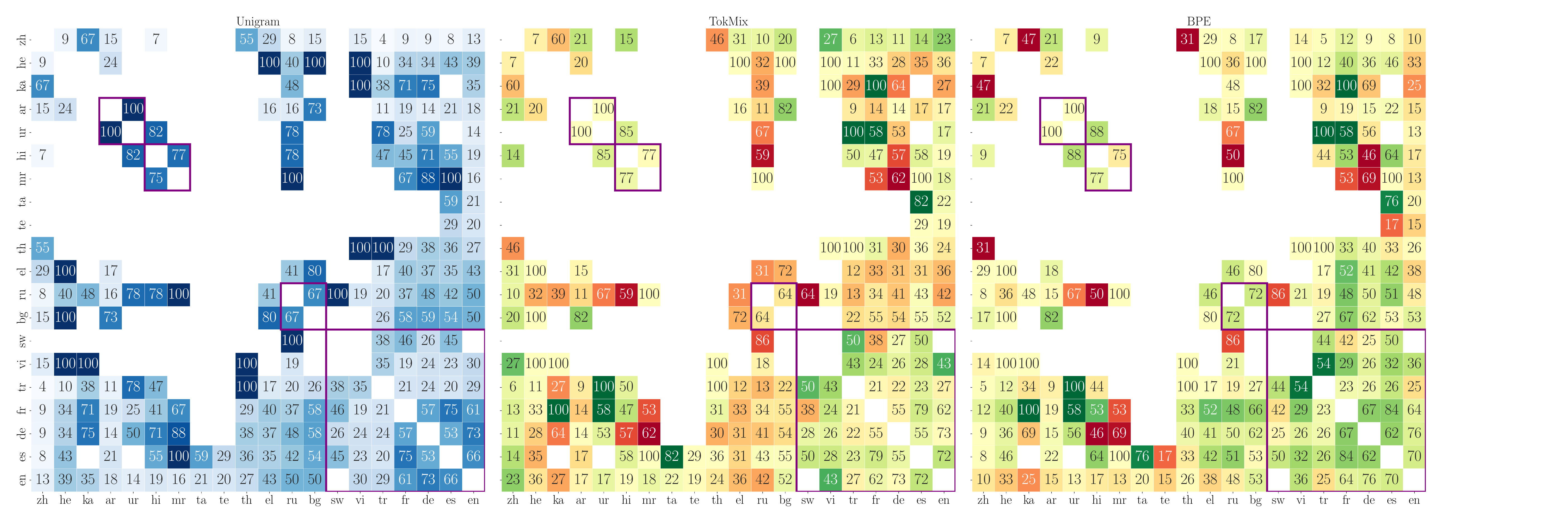}
    
        \caption{Sentence Retrieval}
        \label{fig:tatoeba_transfer_20l}
    \end{subfigure}
    
    \caption{Cross-lingual transfer for the sentence-level tasks for 20 languages. The absolute values are presented for the Unigram tokenizer. For other tokenization methods, we show  the difference from the unigram algorithm.}
\end{figure*}

  \pdfoutput=1

\begin{figure*}[tb!]
    \centering
    \begin{subfigure}[b]{1.0\textwidth}
        \centering
        \includegraphics[width=1.\linewidth]{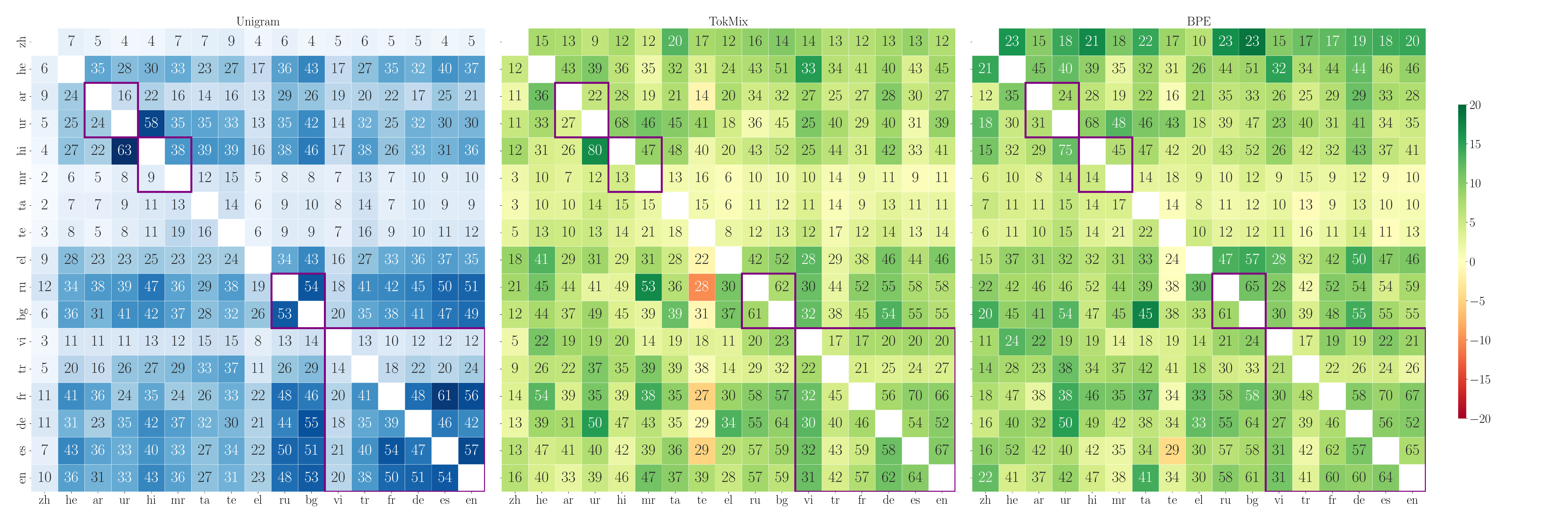}
    
        \caption{Part of Speech Tagging}
        \label{fig:pos_transfer_20l}
    \end{subfigure}

    \begin{subfigure}[b]{1.0\textwidth}

    \centering
    \includegraphics[width=1.\linewidth]{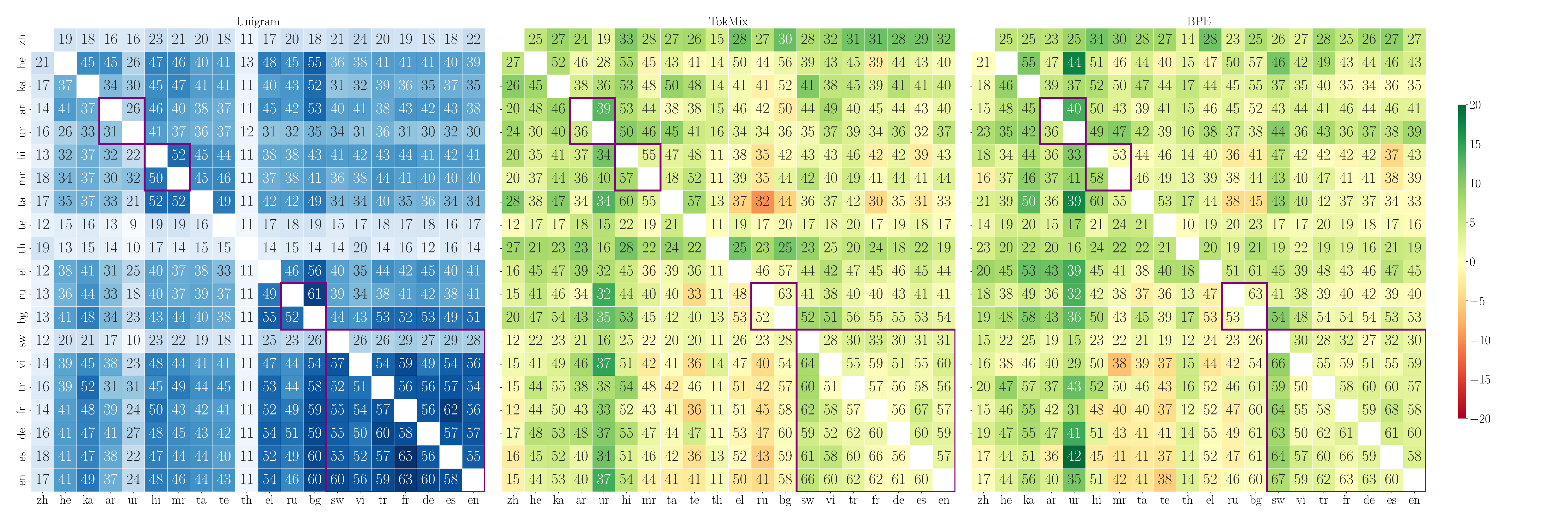}

    \caption{Named Entity Recognition}
    \label{fig:ner_transfer_20l}
    \end{subfigure}
    \begin{subfigure}[b]{1.0\linewidth}
        \centering
        \includegraphics[width=1.\linewidth]{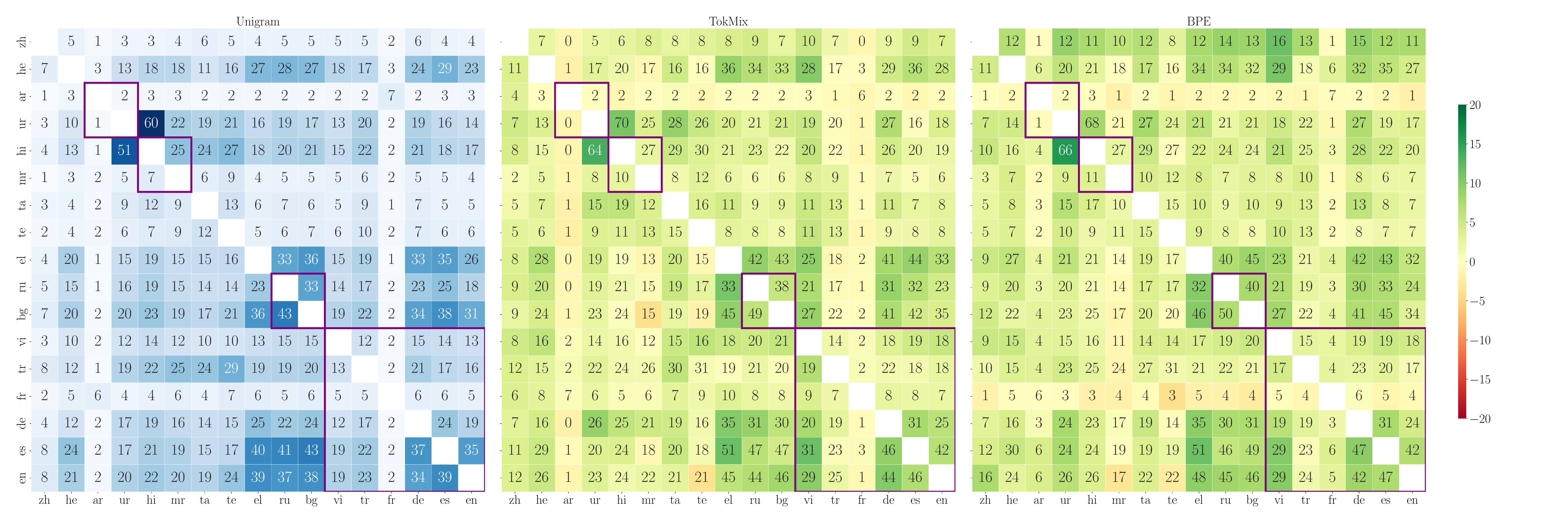}
    
        \caption{Dependency labeling}
        \label{fig:ud_transfer_20l}
    \end{subfigure}
    \caption{Cross-lingual transfer for the token-level tasks on 20 languages. The absolute values are presented for the Unigram tokenizer. For other tokenization methods, we show  the difference from the unigram algorithm.}

\end{figure*}

 The results of experiments for 20 languages: Jensen-Shanon Divergences~\ref{fig:jsd_20l}, and cross-lingual transfers for POS~\ref{fig:pos_transfer_20l}, NER~\ref{fig:ner_transfer_20l}, dependency tree labeling~\ref{fig:ud_transfer_20l}, XNLI~\ref{fig:xnli_transfer_20l}, sentence alignment~\ref{fig:tatoeba_transfer_20l}.

\subsection{Results for All Languages}

We also include detailed results for the in-language experiments along with the proposed tokenizer metrics. In Table~\ref{tab:in_lang}, we present the results for the six languages.

 \subsection{Correlation Analysis}
 \label{sec:correlation_analysis}

\pdfoutput=1
\begin{figure*}[tb!]
    \centering
    \includegraphics[width=1.\linewidth]{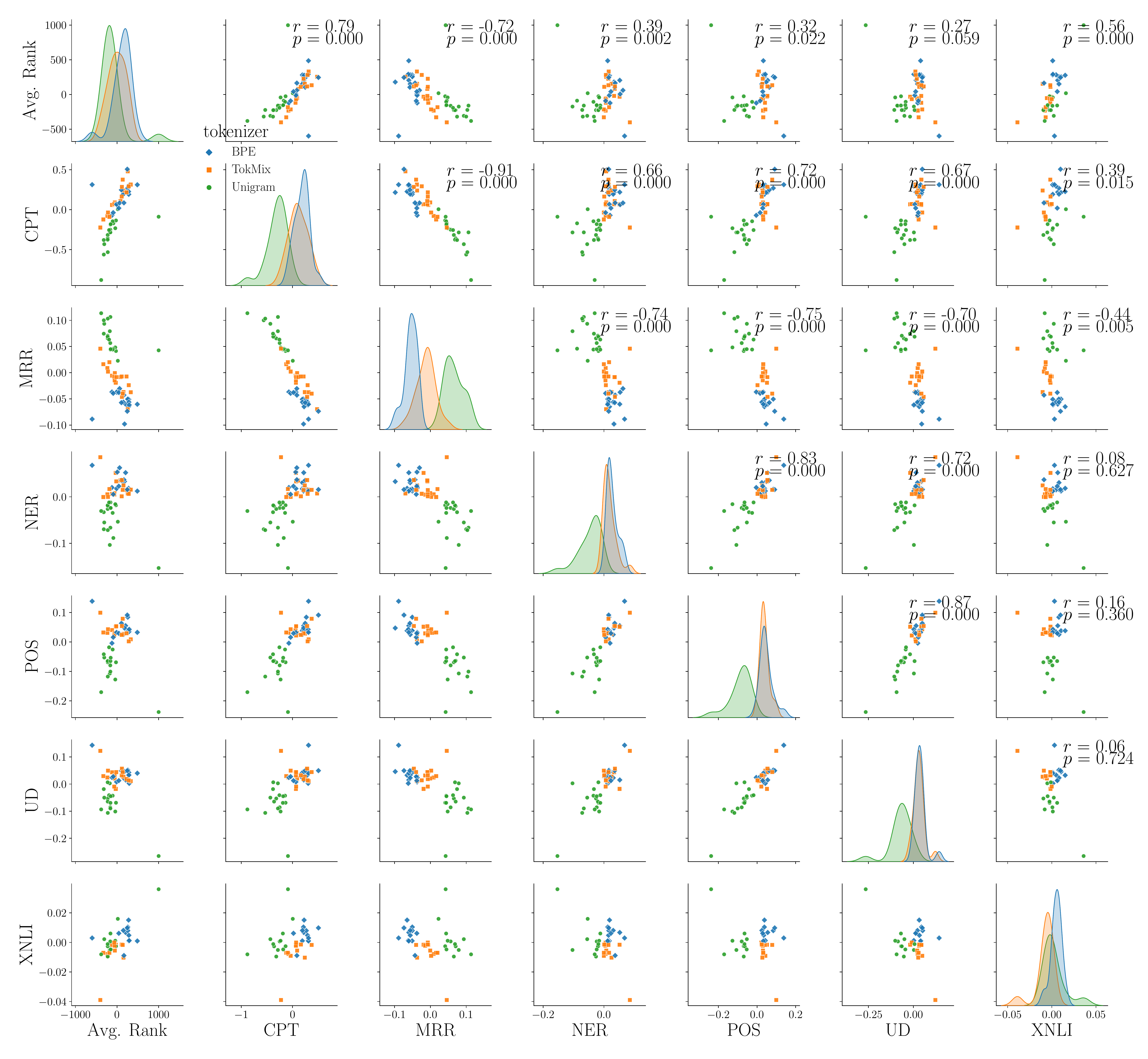}

    \caption{Correlation analysis for pairs of factors: \vo~metrics, language modeling performance (MRR), and downstream tasks.  The diagonal of the figure presents the density of distribution of each feature. The results are grouped by the type of tokenizer applied. Analysis was done in 20 language setting. In the top right corner of each sub-plot, we show Spearman correlation coefficient and associated p-value. }
    \label{fig:correlations_20L}
\end{figure*}

 We present paired correlation plots for in-language metrics in Figure~\ref{fig:correlations_20L}. We use the results from 20 language settings to increase the number of observations. In this analysis, we focus on the differences between the tokenization methods and want to marginalize the language-specific features (such as the pre-training and fine-tuning data size or the model's preference for Indo-European languages). Therefore, for \va{} measures (AR, CPT) and downstream tasks, we subtract the mean for each language. For \vo{} measure (JSD) and transfer values, we subtract the mean value for each pair of languages. In both cases, means are computed across all tokenizers. We present Spearman's correlation coefficient and associated p-value.

\begin{table*}
\centering
\small
\begin{tabular}{lllllllll}
\toprule
     &        &                ar &                tr &                zh &                el &                es &                en &               All \\
metric & tokenizer &                   &                   &                   &                   &                   &                   &                   \\
\midrule
\multirowcell{4}[0pt][l]{\textbf{V. Allocation} \\ (AR)} & Unigram &              2129 &              2719 &              5919 &              2070 &              1439 &              1513 &              2042 \\
     & BPE &              2972 &              3226 &              4294 &              2907 &              2220 &              2143 &              2193 \\
     & NoOverlap &              2537 &              2653 &              2090 &              2065 &              1661 &              1597 &              1829 \\
     & TokMix &              3485 &              4167 &              3961 &              2639 &              1999 &              1898 &              2198 \\
\cline{1-9}
\multirowcell{4}[0pt][l]{\textbf{V. Allocation} \\ (CPT)} & Unigram &              3.16 &              4.01 &              1.84 &               3.5 &              3.88 &              3.91 &              3.17 \\
     & BPE &               3.7 &              4.19 &              2.03 &              3.97 &              4.34 &              4.22 &              4.47 \\
     & NoOverlap &              3.53 &              4.19 &              1.56 &              3.81 &              4.15 &              4.15 &              3.16 \\
     & TokMix &               3.7 &              4.45 &              1.73 &               3.9 &              4.24 &              4.18 &              3.34 \\
\cline{1-9}
\multirowcell{4}[0pt][l]{\textbf{MLM} \\ (MRR)} & Unigram &              36.0 &              36.0 &              34.2 &              46.3 &              49.7 &              49.6 &              42.0 \\
     & BPE &              28.7 &              33.6 &              28.6 &              38.6 &              43.1 &              41.0 &              35.6 \\
     & NoOverlap &              38.1 &              39.6 &              41.4 &              42.8 &              47.5 &              46.6 &              42.7 \\
     & TokMix &              31.5 &              30.6 &              38.2 &              41.2 &              45.3 &              45.6 &              38.7 \\
\cline{1-9}
\multirowcell{4}[0pt][l]{\textbf{NER} \\ (F1)} & Unigram &  66.4 $_{\pm0.1}$ &  73.0 $_{\pm0.1}$ &  35.1 $_{\pm0.1}$ &  68.0 $_{\pm0.1}$ &  68.0 $_{\pm0.1}$ &  66.1 $_{\pm0.2}$ &  62.8 $_{\pm0.1}$ \\
     & BPE &  76.1 $_{\pm0.0}$ &  76.7 $_{\pm0.0}$ &  54.2 $_{\pm0.1}$ &  70.3 $_{\pm0.1}$ &  75.2 $_{\pm0.1}$ &  70.0 $_{\pm0.0}$ &  70.4 $_{\pm0.1}$ \\
     & NoOverlap &  76.5 $_{\pm0.1}$ &  72.8 $_{\pm0.0}$ &  58.4 $_{\pm0.1}$ &  69.6 $_{\pm0.1}$ &  71.6 $_{\pm0.1}$ &  67.3 $_{\pm0.1}$ &  69.4 $_{\pm0.1}$ \\
     & TokMix &  76.6 $_{\pm0.1}$ &  76.2 $_{\pm0.1}$ &  56.1 $_{\pm0.0}$ &  70.1 $_{\pm0.1}$ &  74.3 $_{\pm0.1}$ &  68.1 $_{\pm0.1}$ &  70.2 $_{\pm0.1}$ \\
\cline{1-9}
\multirowcell{4}[0pt][l]{\textbf{POS} \\ (F1)} & Unigram &  54.8 $_{\pm0.1}$ &  46.9 $_{\pm0.2}$ &  29.3 $_{\pm0.1}$ &  52.9 $_{\pm0.3}$ &  76.5 $_{\pm0.2}$ &  81.9 $_{\pm0.1}$ &  57.1 $_{\pm0.2}$ \\
     & BPE &  66.7 $_{\pm0.1}$ &  52.1 $_{\pm0.1}$ &  62.2 $_{\pm0.0}$ &  63.4 $_{\pm0.1}$ &  81.7 $_{\pm0.4}$ &  87.4 $_{\pm0.1}$ &  68.9 $_{\pm0.2}$ \\
     & NoOverlap &  66.5 $_{\pm0.1}$ &  52.5 $_{\pm0.2}$ &  60.6 $_{\pm0.1}$ &  67.5 $_{\pm0.1}$ &  81.3 $_{\pm0.6}$ &  86.7 $_{\pm0.1}$ &  69.2 $_{\pm0.2}$ \\
     & TokMix &  66.0 $_{\pm0.1}$ &  52.1 $_{\pm0.2}$ &  56.2 $_{\pm0.0}$ &  61.7 $_{\pm0.2}$ &  81.3 $_{\pm0.2}$ &  86.3 $_{\pm0.1}$ &  67.3 $_{\pm0.1}$ \\
\cline{1-9}
\multirowcell{4}[0pt][l]{\textbf{Dep. labeling} \\ (F1)} & Unigram &  13.5 $_{\pm0.6}$ &  58.6 $_{\pm0.8}$ &  20.7 $_{\pm0.1}$ &  58.4 $_{\pm0.4}$ &  71.9 $_{\pm0.1}$ &  65.7 $_{\pm0.2}$ &  48.1 $_{\pm0.4}$ \\
     & BPE &  13.8 $_{\pm0.0}$ &  63.7 $_{\pm1.2}$ &  59.5 $_{\pm0.1}$ &  68.2 $_{\pm0.8}$ &  77.0 $_{\pm0.2}$ &  70.3 $_{\pm0.4}$ &  58.7 $_{\pm0.4}$ \\
     & NoOverlap &  13.2 $_{\pm0.0}$ &  65.0 $_{\pm0.5}$ &  60.5 $_{\pm0.2}$ &  67.7 $_{\pm0.2}$ &  77.1 $_{\pm0.3}$ &  69.2 $_{\pm0.3}$ &  58.8 $_{\pm0.3}$ \\
     & TokMix &  14.1 $_{\pm0.0}$ &  62.9 $_{\pm1.2}$ &  53.8 $_{\pm0.1}$ &  67.3 $_{\pm0.5}$ &  76.5 $_{\pm0.1}$ &  69.1 $_{\pm0.2}$ &  57.3 $_{\pm0.4}$ \\
\cline{1-9}
\multirowcell{4}[0pt][l]{\textbf{NLI} \\ (Acc)} & Unigram &  52.5 $_{\pm0.3}$ &  52.9 $_{\pm0.3}$ &  47.5 $_{\pm1.4}$ &  55.0 $_{\pm0.2}$ &  55.3 $_{\pm0.3}$ &  57.4 $_{\pm0.5}$ &  53.4 $_{\pm0.5}$ \\
     & BPE &  52.2 $_{\pm0.3}$ &  53.6 $_{\pm0.5}$ &  45.2 $_{\pm0.4}$ &  55.6 $_{\pm0.3}$ &  55.7 $_{\pm0.2}$ &  57.8 $_{\pm0.2}$ &  53.3 $_{\pm0.3}$ \\
     & NoOverlap &  52.9 $_{\pm0.7}$ &  54.0 $_{\pm0.2}$ &  44.0 $_{\pm0.8}$ &  54.8 $_{\pm0.1}$ &  54.9 $_{\pm0.3}$ &  57.3 $_{\pm0.3}$ &  53.0 $_{\pm0.4}$ \\
     & TokMix &  52.0 $_{\pm0.2}$ &  53.6 $_{\pm0.5}$ &  46.2 $_{\pm1.0}$ &  55.4 $_{\pm0.3}$ &  55.3 $_{\pm0.1}$ &  57.5 $_{\pm0.2}$ &  53.3 $_{\pm0.4}$ \\
\bottomrule
\end{tabular}
\caption{Results of evaluation for in-language properties and tasks for six diverse languages. We observe significant changes for different tokenization methods. The results for MRR, POS, NER, XNLI are in percent. For the downstream task, we show average and standard deviations computed for five runs of probing.}
\label{tab:in_lang}
\end{table*}

\end{document}